\newcommand{\longversion}[1]{#1}
\newcommand{\shortversion}[1]{}

\documentclass[sigconf]{aamas} 

\usepackage{balance} 
\usepackage{times} 
\usepackage{helvet} 
\usepackage{courier} 

\usepackage{amsthm}
\usepackage{caption}
\usepackage{subcaption}
\usepackage{amsmath,amsfonts}
\usepackage[utf8]{inputenc} 
\usepackage[T1]{fontenc}    
\usepackage{url}            
\usepackage{booktabs}       
\usepackage{amsfonts}       
\usepackage{nicefrac}       
\usepackage{microtype}      
\usepackage{xcolor}         
\usepackage{graphicx}
\usepackage{caption}
\usepackage[linesnumbered,ruled,vlined]{algorithm2e}
\usepackage{soul}
\usepackage{siunitx}
\newtheorem{problem}{Problem}

\usepackage{float}
\floatstyle{plaintop}
\restylefloat{table}

\usepackage[mathscr]{euscript}
\let\euscr\mathscr \let\mathscr\relax
\usepackage[scr]{rsfso}

\newcommand\numberthis{\addtocounter{equation}{1}\tag{\theequation}}
\theoremstyle{plain}
\newtheorem{theorem}{Theorem}
\theoremstyle{definition}
\newtheorem{defn}{Definition}

\setcopyright{ifaamas}
\acmConference[AAMAS '24]{Proc.\@ of the 23rd International Conference
on Autonomous Agents and Multiagent Systems (AAMAS 2024)}{May 6 -- 10, 2024}
{Auckland, New Zealand}{N.~Alechina, V.~Dignum, M.~Dastani, J.S.~Sichman (eds.)}
\copyrightyear{2024}
\acmYear{2024}
\acmDOI{}
\acmPrice{}
\acmISBN{}

\acmSubmissionID{<<518>>}

\title{Frugal Actor-Critic:  Sample Efficient Off-Policy Deep Reinforcement Learning Using Unique Experiences}

\author{Nikhil Kumar Singh, Indranil Saha}
\affiliation{
  \institution{Indian Institute of Technology, Kanpur}
  \city{Kanpur}
  \country{India}}
\email{{nksingh,isaha}@cse.iitk.ac.in}

\begin{abstract}
Efficient utilization of the replay buffer plays a significant role in the off-policy actor-critic reinforcement learning (RL) algorithms used for model-free control policy synthesis for complex dynamical systems. 
We propose a method for achieving sample efficiency, which focuses on selecting unique samples and adding them to the replay buffer during the exploration with the goal of reducing the buffer size and maintaining the independent and identically distributed (IID) nature of the samples.
Our method is based on selecting an important subset of the set of state variables from the experiences encountered during the initial phase of random exploration, partitioning the state space into a set of abstract states based on the selected important state variables, and finally selecting the experiences with unique state-reward combination by using a kernel density estimator.
We formally prove that the off-policy actor-critic algorithm incorporating the proposed method for unique experience accumulation converges faster than the vanilla off-policy actor-critic algorithm.
Furthermore, we evaluate our method by comparing it with two state-of-the-art actor-critic RL algorithms on several continuous control benchmarks available in the Gym environment.
Experimental results demonstrate that our method achieves a significant reduction in the size of the replay buffer for all the benchmarks while achieving either faster convergent or better reward accumulation compared to the baseline algorithms.
\end{abstract}

\keywords{Sample efficiency, Reinforcement learning, Control policy synthesis, Off-policy actor-critic algorithm}

\makeatletter
\gdef\@copyrightpermission{
	\begin{minipage}{0.3\columnwidth}
\href{https://creativecommons.org/licenses/by/4.0/}{\includegraphics[width=0.90\textwidth]{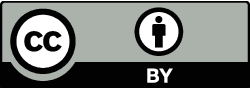}}
	\end{minipage}\hfill
	\begin{minipage}{0.7\columnwidth}	\href{https://creativecommons.org/licenses/by/4.0/}{This work is licensed under a Creative Commons Attribution International 4.0 License.}
	\end{minipage}
	\vspace{5pt}
}
\makeatother

\begin{document}


\pagestyle{fancy}
\fancyhead{}


\maketitle

\section{Introduction}
\label{sec-intro}

The off-policy actor-critic reinforcement learning framework has proven to be a powerful mechanism for solving the model-free control policy synthesis problem for complex dynamical systems with continuous state and action space. 
Experience replay plays a crucial role in the off-policy actor-critic reinforcement learning algorithms. 
The experiences in the form of a tuple containing the current state, the applied action, the next state, and the acquired reward are collected through off-policy exploration and stored in the replay buffer. 
These experiences are replayed to train the critic using optimization algorithms such as stochastic gradient descent (SGD)~\cite{Bottou07,Bottou18}, which in turn guides the training of the actor to improve the target policy.
The performance of the target policy thus depends significantly on the quality of the experiences available in the replay buffer.

To ensure the high quality of the experiences based on which the critic is trained, researchers have proposed several methodologies~\cite{per,lahire22,sinha22} that focus on prioritizing the experiences based on various heuristics. 
However, as the replay buffer size is generally very large to ensure that the samples are independent and identically distributed (IID), prioritizing the samples incurs a significant computational overhead.
In this paper, we rather explore an alternative strategy to ensure the quality of the experiences on which the critic is trained. We aim to devise a mechanism that can identify unique experiences and store them in the replay buffer when they provide knowledge different from what is already available with the experiences present in the replay buffer. Maintaining unique experiences in the replay buffer has two major advantages.  
First, having unique experiences in the replay buffer ensures that the samples are IID, which is essential for the stochastic gradient descent (SGD) optimization algorithm used in training the critic to converge faster by not suffering from instability caused by high estimation variance~\cite{Needell14,Zhao15} and without the necessity of employing expensive methods for prioritizing the samples within the reply buffer. 
Second, it helps in reducing the memory requirement, which is crucial for deploying actor-critic RL algorithms in memory-constrained embedded systems~\cite{Slimani19,Sahoo21}.

In this paper, we present an algorithm named Frugal Actor-Critic (FAC) that strives to maintain unique experiences in the replay buffer. Ideally, the replay buffer should contain those experiences that provide unique learning to the agent. From a state, it may be possible to reach some other state and acquire the same reward by applying different actions. However, as the goal of the agent is to maximize the accumulated reward, it does not need to learn all the actions. Rather, it is enough to learn one of those actions that lead to a specific reward from a state while moving to another state. Thus, FAC attempts to identify experiences based on unique state-reward combinations.

FAC employs a state-space partitioning technique based on the discretization of the state variables to maintain unique experiences in the replay buffer. However, a complex dynamical system may have a high-dimensional state vector with correlated components. Thus, state-space partitioning based on all the state variables becomes computationally expensive. FAC employs a matrix decomposition based technique that uses the experiences encountered during the initial phase of random exploration to identify the most important and independent state variables that characterize the behavior of the dynamical system. Subsequently, the state space is partitioned based on the equivalence of the values assumed by the selected state variables. The goal of the FAC algorithm is to store experiences that are from different partitions of the state space.
However, naively discarding samples whose state values are similar to those present in the replay buffer might lead to degradation in learning (since samples with similar states can have different rewards)  
and hence synthesis of sub-optimal control policy. Since learning depends on reward maximization, the control policy needs the knowledge of different rewards achieved for a given state (by taking different actions). 
To accommodate different rewards possible for similar states, we introduce a mechanism for estimating the density of a reward based on a kernel density estimator~\cite{Terrell92}. This mechanism aims to identify those experiences for a given state partition whose corresponding rewards are under-represented in the replay buffer.

The computational overhead of FAC is negligible, and it obviates the necessity of employing computationally demanding sample prioritization methods to improve the performance of the replay buffer based RL algorithms.
We theoretically prove that the FAC algorithm is superior to the general off-policy actor-critic algorithms in storing IID samples in the replay buffer as well as in convergence. 
Moreover, We evaluate FAC experimentally by comparing it with two state-of-the-art actor-critic RL algorithms SAC~\cite{haarnoja18} and TD3~\cite{fujimoto18} on several continuous control benchmarks available in the Gym environment~\cite{gym}. Experimental results demonstrate that our method achieves a significant reduction in the size of the replay buffer for all the benchmarks while achieving either faster convergent or superior accumulated reward compared to the baseline algorithms. 
We also compare FAC with the state-of-the-art sample prioritization technique LABER~\cite{lahire22}, demonstrating that FAC consistently outperforms LABER in terms of convergence and accumulated reward. 
To the best of our knowledge, this is the first work that attempts to control the entries in the replay buffer to improve the performance of the off-policy actor-critic RL algorithms.


\section{Problem}

\subsection{Preliminaries}

\paragraph{Reinforcement Learning.}

To solve a control policy synthesis problem using reinforcement learning, we represent the model-free environment $\mathcal{M}$  as a tuple $\langle S,A,\mathcal{T},R\rangle$,
where $S$ denotes the set of continuous states $s\in\mathbb{R}^p$ of the system and $A$ denotes the set of continuous control actions $a\in\mathbb{R}^q$. The function $\mathcal{T}:\mathbb{R}^{p\times q\times p}\rightarrow[0,\infty)$ represents the unknown dynamics of the system, i.e., the probability density of transitioning to the next state given the current state and action. 
A transition (or sample) $x$ at time $t$ is denoted by $( s_t,a_t,r_t,s_{t+1} )$, where $s_t\in S$ is a state, $a_t\in A$ is an action, $r_t \in \mathbb{R}$ is the reward obtained by the agent by performing action $a_t$ at state $s_t$, and $s_{t+1} \in S$ is the next state.
We denote by $R$ the reward function for $\mathcal{M}$ such that $R(\omega)=\sum_{t=0}^{T} \gamma^t\cdot r_t$, where $\omega=(x_0,\ldots,x_T)$  
is the trace generated by a policy, and $\gamma$ is the discount factor.
To find the optimal policy $\pi^*: S\rightarrow A$, we define a parameterized policy $\pi_{\theta}(a|s)$ as the probability of choosing an action $a$ given state $s$ corresponding to parameter $\theta$, i.e.,
\begin{equation}
\pi_{\theta}(a|s) = \mathbb{P}[a|s;\theta].
\end{equation}

We define the cost function (reward) associated with the policy parameter $\theta$ as follows:
\begin{equation}
\label{eqpd}
J(\theta) =  \underset{\omega \sim \pi_{\theta}}{\mathbb{E}} \ [R(\omega)],
\end{equation}
where $\underset{\omega \sim \pi_{\theta}}{\mathbb{E}}$ denotes the expectation operation over all traces $\omega$ generated by using the policy $\pi_{\theta}$.

Our goal is to learn a policy that maximizes the reward. 
Mathematically,
\begin{equation}
\theta^* = \underset{\theta}{\arg\max} \ 
    \  J(\theta).
\end{equation}

Hence, our optimal policy would be the one corresponding to $\theta^*$, i.e., $\pi_{\theta^*}$.

\paragraph{Deep RL Algorithms with Experience Replay.}

The direct way to find the optimal policy is via policy gradient (PG) algorithms. However, the PG algorithms are not sample-efficient and are unstable for continuous control problems. Actor-critic algorithms~\cite{Konda03}, on the other hand, are more sample-efficient as they use Q-function approximation (critic) instead of sample reward return. 

The gradient for the Actor (policy) can be approximated as follows: 
\begin{equation}
\label{grad}
    \nabla_{\theta}J(\theta) = 
   {\mathbb{E}_{\pi_{\theta}}}  \
    \ [ \sum_{t=0}^{T}
    \ \nabla_{\theta} \ log \ \pi_{\theta}(a_t|s_t) \cdot Q(s_t,a_t)]. 
\end{equation}
The function $Q(s_t,a_t)$ is given as:
\begin{equation}
\label{qfcn}
Q(s_t,u_t)= \mathbb{E}_{s_{t+1} \sim \mathbb{P}(.|s_t,u_t)}    [ r_t +\gamma*Q(s_{t+1},u_{t+1})].
\end{equation}

The policy gradient, given by
Equation~\eqref{grad}, is used to find the optimal policy. 
The critic, on the other hand, uses an approximation to the action-value function to estimate the function $Q(s_t,a_t)$.
This approximation function is learned using transitions stored in a \emph{replay buffer} 
which are collected under the current policy. 
Experience replay, i.e., storing the transitions in a replay buffer $\mathcal{R}$, plays a vital role in actor-critic based algorithms. It enables efficient learning by reusing the experiences (transitions) multiple times. Moreover, as the replay buffer samples are more IID compared to that in the case of on-policy, the learning is more stable due to better convergence.

We use the term general actor-critic (GAC) algorithms to denote the general class of off-policy actor-critic algorithms that use samples from the replay buffer for learning.




\subsection{Problem Definition}


Before we present the problem addressed in the paper formally, let us motivate the need to maintain unique samples in the replay buffer while synthesizing a control policy using the classical example of controlling an inverted pendulum to maintain its upright position.
By default, a pendulum freely hangs in the downward position, and the goal is to balance it vertically upward.
During the exploration phase, it spends much time swinging around the downward position towards the left and right. Consequently, the replay buffer is populated with samples mostly from a small region of the state space (around the downward position). In the learning phase, this behavior leads to oversampling from a specific subset of the experiences and under-sampling from the rest.
This phenomenon leads to the less frequent replay of critical and rare experiences (for example, those corresponding to the pendulum's upright position), leading to the delay in learning the control policy. 

The above discussion emphasizes the importance of 
collecting the experience from all regions of the state space uniformly. In general, maintaining unique experiences in the replay buffer is essential to maintain the IID characteristics of the samples, which is crucial for the superior performance of the optimization algorithms used for learning. Moreover, it also helps in optimizing the size of the replay buffer, which, though may not pose any challenge for a simple system like the pendulum, may be a bottleneck for large and complex systems such as a humanoid robot.
Keeping this in mind, we now define the problem formally. 

\begin{defn}[Equivalent Experience]
\label{eqv-def}
Let us consider a transition $x= \langle s_t,a_t,r_t,s_{t+1} \rangle$. We define a transition $x$ as an (approximate) \textit{equivalent} of $x'$ if following condition hold:
\begin{itemize}
    \item \texttt{state}\,($x$) and \texttt{state}\,($x'$) are in close proximity in some metric space, where  \texttt{state}\,($x$) = $s_t$.
    \item \texttt{reward}\,($x$) and \texttt{reward}\,($x'$) are in close proximity in some metric space, where \texttt{reward}\,($x$) = $r_t$. 
\end{itemize}
Note that the two metric spaces for state and reward can be different. A transition $x$ is \textit{unique} w.r.t. a replay buffer $\mathcal{R}$  if $\mathcal{R}$ does not contain any transition $x'$ which is equivalent to $x$. 
\end{defn}    

\begin{problem}[Sample Efficiency] 
\label{prob-def}
Let $\mathcal{M}$ be an open-loop dynamical system with unknown dynamics. Synthesize a control policy $\pi^*$ with replay buffer $\mathcal{R}$ only containing unique samples , so that the expectation of rewards generated by the policy gets maximized.
Mathematically,
    \begin{align*}
    \pi^*= \underset{\pi}{\arg\max} \  \underset{\substack{{\omega \sim \mathcal{R}^{\pi}} \\ \mathcal{R}^{\pi} \sim \mathtt{sim }(\pi,\mathcal{M})}}{\mathbb{E}} \ [R(\omega)]
    \\
    \text{s.t.~ $\mathcal{R}^{\pi}$ contains unique samples}.\numberthis
    \label{prob}
\end{align*}
\end{problem}

\section{The FAC Algorithm}
 \label{method}
The FAC Algorithm is presented in Algorithm~\ref{algo1}. FAC takes the model of the environment $\mathcal{M}$ and the number of time-steps  $T$ as input and returns the optimal control policy $\pi^*$. For this purpose, 
we start with a random policy (line~\ref{l0}) and collect transitions  
(rollouts) using it (line~\ref{l1}). Then, we find the important dimensions of the state-space from the collected rollouts using $\mathtt{find\_important\_dimensions}$ method (line~\ref{l2}). Along these important dimensions, the state-space is partitioned using $\mathtt{partition\_state\_space}$ method (line~\ref{l3}) to create a set of abstract states. Then, these abstract states are used to identify unique samples using $\mathtt{select\_samples\_for\_insertion}$ function to find the optimal policy (line~\ref{l4}). 
We now present each of these functions in detail.

\begin{algorithm}[t]
\DontPrintSemicolon 
\SetKwProg{myproc}{procedure}{}{}

\myproc{$\mathtt{frugal\_actor\_critic}$($\mathcal{M}$, $T$)}{
$\pi_{\theta} \gets \mathtt{initialize\_policy}()$\;\label{l0}
$\Omega \gets \mathtt{initial\_rollout } (\mathcal{M,\pi_{\theta}})$\; \label{l1}

$\kappa \gets \mathtt{find\_important\_dimensions }(\Omega) $\; \label{l2}

$\alpha \gets \mathtt{partition\_state\_space }(\mathcal{M},\kappa,\mu) $\; \label{l3}
$\pi^* \gets \mathtt{learn }(\mathcal{M},\pi_{\theta},\alpha,T)$\; \label{l4}

\Return{$\pi^*$}\; \label{l5}
}

\myproc{$\mathtt{find\_important\_dimensions }(\Omega)$}{
$\langle \mathtt{Q}, \mathtt{R}, \mathtt{E} \rangle \gets \mathtt{matrix\_decomposition}(\Omega)$ \; \label{l6}
 $idx \gets \mathtt{index }(|\mathtt{diag}(R)|\geq \nu \cdot |\mathtt{diag}(R)[0]|)$ \; \label{l7}
$\kappa \gets E[\mathtt{diag}(R)[idx]]$ \; \label{l8}

\Return{$\kappa$}\;  \label{l9}
}

\myproc{$\mathtt{learn }${ ($\mathcal{M},\pi_{\theta},\alpha$,$T$)}}{
$\mathcal{R} \gets \emptyset$; \, 
$\phi \gets \{\}$; \,
$s_0 \sim \rho^{\pi}$\;

\While{$t~<~T$}{\label{l10}

$a_t \gets \pi_{\theta}(s_t)$\; \label{l11}
$\langle s_{t+1},r_t \rangle \gets \mathtt{sim }(\mathcal{M},s_t,a_t)$\;\label{l12}

$\mathcal{R} \gets \mathtt{select\_samples\_for\_insertion }(\phi,\alpha,s_t,r_t)$\; \label{l13}

\texttt{Sample~b~transitions~$\langle s_i,a_i,r_i,s_{i+1}\rangle$ from~$\mathcal{R}$}\;\label{l14}
$\delta_i \gets r_i + \gamma\cdot \underset{a}{\mathtt{max}}~Q(s_{i+1},a;\theta) - Q(s_i,a_i;\theta)$\; \label{l15}
$\theta_{k+1} \gets \theta_{k} + lr* \sum_{i=1}^{b}\delta_i\cdot \nabla_{\theta} Q(s_i,a_i;\theta)$\; \label{upd}
$t \gets t + 1$\;
}
\Return{$\pi_{\theta}^*$}\;  \label{l16}
}

\myproc{$\mathtt{select\_samples\_for\_insertion }(\phi,\alpha,s_t,r_t)$}{

$\euscr{R}(r_t,s_t) \gets \int_{r_t-\beta}^{r_t+\beta}\rho_K(y,s_t)\; dy $\;\label{l17}

$ \epsilon^s = \frac{\epsilon}{\mathtt{exp }(\frac{|\phi [\alpha(s)]|}{\eta})}$\;\label{l175}

\If{$\euscr{R}(r_t,s_t) <\epsilon^s$}{\label{l18}
 $ \mathcal{R} \gets \mathcal{R} ~ \cup \langle s_t,a_t,s_{t+1},r_t \rangle $\;\label{l19}
$\phi[\alpha(s_t)] \gets \phi[\alpha(s_t)]~ \cup~ \{r_t\} $\;\label{l20}
}
\Else{
 $\mathtt{discard }(\langle s_t,a_t,s_{t+1},r_t \rangle)$\;\label{l21}
}

\Return{$\mathcal{R}$}\;  \label{l22}
}

\caption{{\sc Frugal Actor-Critic}}
\label{algo1}
\end{algorithm}

\subsection{Finding Significant State Dimensions}
\label{md}

The FAC algorithm identifies unique experiences based on the abstract states created through the partition of the state space. However, for high-dimensional complex nonlinear systems, naive partitioning of the state space leads to a large number of abstract states, hindering the scalability of the algorithm. Several states of a high-dimensional system have insignificant impacts on the overall behavior of the system. Moreover, as some of the state dimensions might be correlated, considering all the state dimensions unnecessarily introduce several abstract states that are not reachable by the system. Thus, as the first step (Lines~\ref{l0}-\ref{l2}), FAC finds a subset of independent state variables that captures the unique behaviors of the system. 

FAC collects the trace $\Omega = (s_0, s_1, \ldots,s_{\euscr{T}-1})$, $s_i$ denoting the state at the $i$-th step, based on the random policy $\pi_\theta$ used in the initial rollout process before the learning starts at the $\euscr{T}$-th step. These traces are now used to extract the subset of significant state dimensions $\kappa$.
This is reasonable as the initial policy is completely random, and hence the data collected is sufficiently exhaustive to approximate the overall behavior. One could imagine that Principal Component Analysis (PCA)~\cite{WOLD87} is suitable for implementing this step. However, though PCA is capable of providing a low-dimensional representation of the original high-dimensional data, it is incapable of identifying the subset of state dimensions that contributes the most to the low-dimensional representation. 
Thus, we rather employ the rank-revealing QR decomposition~\cite{CHAN87}, which gives us the state dimensions (i.e., features) sorted according to their singular values.
This enables us to discard the less critical state dimensions containing minor information (i.e., low variance) or contain information that is already present in some other dimension (i.e., high correlation).

To apply QR decomposition on the initially collected system trace $\Omega$, the function $\mathtt{find\_important\_dimensions}$ creates a matrix $[s_0^\mathtt{T} \, s_1^\mathtt{T} \, \ldots \, s_{\euscr{T}-1}^\mathtt{T}]^\mathtt{T}$ with time-step as rows and state dimension as columns.
The function $\mathtt{matrix\_decomposition }$ refers to the permuted-QR decomposition such that \mbox{$\Omega[:,E]=Q\cdot R$}, where $E$ is the permutation matrix. The function $\mathtt{index}$ fetches the index of columns (state dimensions) that satisfies the input constraints, i.e., features whose singular value is larger than $\nu$ times the largest singular value among all features.
Here $\nu \in \mathbb{R}$ is a hyperparameter having a value between $0$ and $1$.

\subsection{State-Space Partitioning}
\label{part}

Once the important state dimensions $\kappa$ are decided, FAC partitions the state space along those dimensions using the $\mathtt{partition\_state\_}$ $\mathtt{space}$ function (line~\ref{l3}).
It uses a vector $\mu$ of length $|\kappa|$ whose element $\mu_i$ decides the granularity of the $i$-th important state, i.e., the range of values of the $i$-th important state is divided into $\mu_i$ partitions.
Overall, the state space partitioning results in a non-uniform multi-dimensional grid structure, leading to an abstract state-space as $\mathcal{A}=\langle a_1,\ldots,a_{v} \rangle$ where $a_i \in\mathbb{R}^{\kappa}$.
The hyperparameter $\mu$ controls the coarseness of the partition of the abstract state-space --- smaller values for the elements in $\mu$ lead to a larger number of abstract states, i.e., the value of $v$. Note that although we have infinite states in the original state-space, we have a finite ($v$) number of abstract states.

We use $\alpha: S \rightarrow \mathcal{A}$   to denote a state partition function.
For a state $s \in S$, we denote the vector containing the values in $\kappa$ dimensions,  preserving the order as $s^{\kappa}$. The corresponding mapped abstract state $a$ for state $s$ is given as
\begin{align*}
\label{eq:abs}
    a = \underset{\hat{a}\in \mathcal{A}}{\mathtt{arg~min}}\, \lVert s^{\kappa} - \hat{a}\rVert\numberthis.
\end{align*}

\subsection{Density Estimation} 
\label{de}

Once the abstract states are identified, FAC requires a mechanism to identify experiences providing unique rewards for an abstract state.
We create a function $\phi: \mathcal{A}\rightarrow \euscr{P}(\hat{R})$,  where $\euscr{P}(\hat{R})$ represents the power set of the set $\hat{R}$ having all the distinct reward values  present in the replay buffer. 
Here, $\phi$ is a mapping from each abstract state to a set of dissimilar (unique) reward values. We use density estimation to control the insertion of a sample  
$\langle s,a,r,s' \rangle$ into the replay buffer,
i.e.,
we estimate the density of the reward corresponding to the new sample using the set of rewards corresponding to an abstract state. If the new sample's reward density is low (i.e., distinct), then we insert it, else we reject it.

Our density estimation procedure is motivated by the kernel density estimator (KDE)~\cite{Terrell92}. We define the density estimation function for a reward $r$ w.r.t a given state $s$ as follows:
\begin{equation}
\label{kde}
    \rho_K(r,s) = \sum_{i} K(r-\phi[\alpha(s)]_i;h),
\end{equation}
where $i$ refers to the indexes of the different rewards corresponding to $\phi[\alpha(s)]$.
Also, $K$ refers to the kernel function, and $h$ refers to the bandwidth, which acts as a smoothing parameter. 


We define the reward density estimate (RDE) $\euscr{R}$ of a reward sample $r$ as follows:
\begin{equation}
\label{eqp}
    \euscr{R}(r,s) = \int_{r-\beta}^{r+\beta}\rho_K(y,s)\cdot dy,
\end{equation}
where $\beta$ controls the space width around the reward $r$ used to compute reward density.
An experience with state $s$ and reward $r$ is stored in the replay buffer if $\euscr{R}(r,s) \le \epsilon$, where $\epsilon \in (0,1)$ is a hyperparameter.

Generally, for a given abstract state, if we have a high RDE for any reward value (or range), this indicates that we have enough samples in that reward region. Consequently, more samples are not needed in that region. However, we might encounter a scenario where an abstract state has a large number of samples for those many different rewards. Consequently, each sample's probability mass contribution would be much less for the RDE of any reward region, ultimately leading to low RDE for all the reward regions of that abstract state. A low RDE by default would encourage new samples to be added to the replay buffer (since $\euscr{R}(r,s)<\epsilon$), which we want to avoid as we already have enough samples. 
%
%
%
%
To address this issue, we use a dynamic value of $\epsilon$ by scaling it $1/\mathtt{exp }(\frac{|\phi [\alpha(s)]|}{\eta})$ times as follows: 
\begin{equation}
    \epsilon^s = \frac{\epsilon}{\mathtt{exp }(\frac{|\phi [\alpha(s)]|}{\eta})}.
\end{equation}
Here $\eta$ is a hyperparameter that represents a large number to  keep the value of the exponential term closer to 1 in general and to make it attain a higher value only for an extremely large number of samples. 

In Algorithm~\ref{algo1}, RDE is computed at line~\ref{l17}. In case of low RDE (line~\ref{l18}), the transition tuple is inserted into replay buffer $\mathcal{R}$ (line~\ref{l19}) and $\phi$ is updated (line~\ref{l20}). In case the reward density is high, i.e., there are already a sufficient number of samples in the region, the transition is discarded (line~\ref{l21}).

\subsection{Learning from Replay Buffers.} 
\label{learn}
In Algorithm~\ref{algo1}, the learning continues for $T$ time-steps (line~\ref{l10}). Each simulation step of the policy under the environment at time $t$ generates the transition tuple $\langle s_t,a_t,r_t,s_{t+1}\rangle$ (line~\ref{l12}). In line~\ref{l13}, the function $\mathtt{select\_samples\_for\_insertion }$ to compute the RDE of the sample at time $t$ is invoked, which in turn adds the sample to the replay buffer or rejects it. Subsequently, a mini-batch of samples is uniformly sampled from the replay buffer (line~\ref{l14}) and is used to update the control policy parameters $\theta$ (line~\ref{upd}).

FAC controls the insertion of samples into the replay buffer, thereby ensuring the uniqueness of samples present there. Practically, this is analogous to a sampling strategy where we pick unique samples. Controlling insertions into the replay buffer rather than sampling from the replay buffer is primarily driven by computational efficiency. Due to this, FAC incurs very little overhead vis-$\grave{a}$-vis the GAC algorithms.
The uniform sampling strategy (UER), which is computationally the least expensive~\cite{Mnih2015}, is used in FAC.

\subsection{Computation Overhead} 
\label{compo}
We now present an analysis of the complexity of the operations used in the FAC algorithm. In Line~\ref{l1}-\ref{l3}, we perform a few additional computations before starting the control policy synthesis. All these operations are performed before the training starts and involve negligible overhead. 
While inserting samples into the replay buffer, we perform some additional computations in Line~\ref{l17} for density estimation for a given sample using the samples in its close vicinity (the number of such samples is fewer due to our insertion strategy). Hence, the time complexity of the insertion operation of FAC remains $\mathcal{O}(1)$, i.e., constant in terms of the number of samples.
Although negligible cost, the additional computations that we perform help us achieve the sampling from the replay buffer in $\mathcal{O}(1)$ time due to uniform sampling strategy. 
Prioritization sampling, on the other hand, is computationally expensive with $\mathcal{O}(\log\, n)$ complexity due to the updation of priorities of the batch samples (via priority queues).
We circumvent this issue by controlling the insertion into the replay buffer (using operations having negligible cost) rather than performing prioritized sampling from the replay buffer.

\section{Theoretical Analysis}



We use the notion of regret bound~\cite{Gordon99,Shwartz12} to compare the convergence of FAC algorithms vis-a-vis the GAC algorithm.
The standard definition of cost function in reinforcement learning (refer Equation~\ref{eqpd}) uses a function of policy parameter $\theta$. In this case, the traces/transitions are implicit as they are generated from a policy with parameter $\theta$ (i.e., $\pi_{\theta}$). However, in the off-policy setting, the transitions used for learning are not directly generated by the policy but sampled from the replay buffer. To take this into account, we denote the cost function as $J(\theta,x)$ instead of $J(\theta)$ where $x$ represents the transitions from replay buffer $\mathcal{R}$.




\begin{theorem}[Convergence]
\label{thm2}
Let $J(\theta,x)$ be an L-smooth convex cost function in policy parameter $\theta$ and $x \in \mathcal{R}$. 
Assume that the gradient $\nabla_{\theta} J(\theta,x)$ has $\sigma^2$-bounded variance for all $\theta$.
Let $\zeta$  represent the number of equivalent samples in a mini-batch of size $b$ ($\zeta\leq b$) for GAC algorithms. 
Then, FAC would converge $\frac{b+\zeta^2+\zeta}{b}$ times faster than the GAC algorithms.
\end{theorem}
%
\shortversion{
\begin{proof}
    Refer~\cite{singh24}.
\end{proof}
}
\longversion{

\begin{proof}
Let $m$ represent the total number of samples in the replay buffer. Let $b$ represent the number of samples in a mini-batch.   Assume, without loss of generality, that $m$ is divisible by $b$. Let $\bar{x}=( x_1,\ldots,x_b)~\in\mathcal{R}^b$ represent mini-batch set of size $b$ sampled from replay buffer $\mathcal{R}$. Let us assume that each mini-batch contains $\zeta$ equivalent samples in expectation. 
We denote the space of control policy parameter $\theta$ as $\Theta$.
Let $\bar{J}: \Theta\times\mathcal{R}^b \rightarrow \mathbb{R}$ be the mean of cost function $J$ over a given mini-batch, 
defined as:
\begin{equation}
\label{th3_1}
\bar{J}(\theta,(x_1,\ldots,x_b)) = \frac{1}{b} \sum_{s=1}^{b}\, J(\theta,x_s).
\end{equation}

The true mean cost function over whole replay buffer $\mathcal{R}$ is given as
\[ \mathcal{J}(\theta) = \mathbb{E}_{x \in \mathcal{R}} \, J(\theta,x)   \]



From Equation~\ref{th3_1}, we compute the gradient of the mini-batch cost function using linearity of gradients, as follows
\[\nabla_{\theta} \bar{J}(\theta,(x_1,\ldots,x_b)) = \frac{1}{b} \sum_{s=1}^{b} \nabla_{\theta} J(\theta,x_s)\]

The euclidean norm of variance of the gradients of mini-batch cost function w.r.t. true mean cost function is given as: 
\begin{align*}
 \Vert \nabla_{\theta} \bar{J}(\theta,\bar{x}) - \nabla \mathcal{J}(\theta) \Vert^2  &=  \Vert \frac{1}{b} \sum_{s=1}^{b} (\nabla_{\theta} J(\theta,x_s) - \nabla \mathcal{J}(\theta) )  \Vert^2 \\
 \begin{split}
  &=   \frac{1}{b^2} \sum_{s=1}^{b} \sum_{s'=1}^{b} \langle\nabla_{\theta} J(\theta,x_s) - \nabla \mathcal{J}(\theta), \\ &\quad \quad  \nabla_{\theta} J(\theta,x_{s'}) - \nabla \mathcal{J}(\theta) \rangle 
  \end{split} \numberthis 
  \label{thm3_3}
\end{align*}

Here $\langle~,\rangle$ represents the dot product.

\paragraph{Case 1: GAC} Now we derive the regret bound in case of GAC algorithm.   
 Observe the unique samples in the Equation~\ref{thm3_3} (refer Definition~\ref{eqv-def}). 
 Note that there are $\zeta$ equivalent samples in each set (mini-batch)
 and due to this we get $\zeta \cdot(\zeta+1)$ less elements. 
 Therefore, $b^2-(b+\zeta^2+\zeta)$  elements in the product vanish, i.e.,
%
\begin{align*}
&\mathbb{E}[ \langle\nabla_{\theta} J(\theta,x_s) - \nabla \mathcal{J}(\theta), \nabla_{\theta} J(\theta,x_{s'}) - \nabla \mathcal{J}(\theta) \rangle]  = \\
&\ \ \ \ \ \ \ \ \ \ \ \ \langle \mathbb{E} [\nabla_{\theta} J(\theta,x_s) - \nabla \mathcal{J}(\theta)], \mathbb{E}[\nabla_{\theta} J(\theta,x_{s'}) - \nabla \mathcal{J}(\theta)] \rangle  = 0.
\end{align*}
%
Moreover, we have $\sigma^2$-bounded gradient variance, i.e.,
\[\mathbb{E}_x[ \Vert \nabla_{\theta} J(\theta,x) - \nabla \mathcal{J}(\theta) \Vert^2  ] \leq \sigma^2.\]
Hence,
\begin{align*}
 \mathbb{E} & \Vert \nabla_{\theta} \bar{J}(\theta,\bar{x}) - \nabla \mathcal{J}(\theta) \Vert^2 \\
 & = \frac{b+\zeta^2+\zeta}{b^2}\cdot \frac{1}{b+\zeta^2+\zeta} \sum_{s=1}^{b+\zeta^2+\zeta} \Vert  (\nabla_{\theta} J(\theta,x_s)  - \nabla \mathcal{J}(\theta) )  \Vert^2 
\\
 &\leq \frac{\sigma^2\cdot (b+\zeta^2+\zeta)}{b^2}
\end{align*}
%
%
The above equation shows that $\nabla_{\theta} \bar{J}(\theta,\bar{x}_j)$ has $\frac{\sigma^2\cdot (b+\zeta^2+\zeta)}{b^2}$ bounded variance and hence the regret bound of $\bar{J}$ is given as
\[ \mathbb{E} [ \sum_{j=1}^{m/b} (\bar{J}(\theta_j,\bar{x}_j) - \bar{J}(\theta^*,\bar{x}_j))]   \leq \psi(\frac{\sigma^2\cdot (b+\zeta^2+\zeta)}{b^2},\frac{m\cdot (b+\zeta^2+\zeta)}{b^2})\]
Using Equation~\ref{th3_1} and multiplying by $b$, we get
\begin{align*}
  \mathbb{E} [ \sum_{j=1}^{m/b} \sum_{i=(j-1)b+1}^{jb} & (J(\theta_j, x_i) - J(\theta^*,x_i))]   \leq   \\  
  & \ \ \ \ \ \ \ \ \ \ \ \ \ \ 
  b\cdot \psi(\frac{\sigma^2\cdot (b+\zeta^2+\zeta)}{b^2},\frac{m\cdot (b+\zeta^2+\zeta)}{b^2}).
\end{align*}

The goal of deriving the regret bounds for the GAC algorithms is to estimate its convergence rate. There are many formulations of the function $\psi$ in literature for such comparison. Here, we use Theorem~2 in~\cite{dekel12} to define $\psi$ as
\[\psi(\sigma^2,m)=2D^2L+2D\sigma\sqrt{m},\]
%
where $L$ refers to L-smoothness of function $J$ and $D=\sqrt{ \frac{\mathtt{max}_{\theta_1,\theta_2 \in \Theta} \lVert \theta_1 - \theta_2 \rVert^2}{2}}$.


Thus, the regret bound in case of GAC algorithms is
\[2bD^2L + 2D\sigma \frac{b+\zeta^2+\zeta}{b} \sqrt{m}. \]

\paragraph{Case 2: FAC} In case of FAC, there are no equivalent samples in the replay buffer $\mathcal{R}$. Therefore the $b^2-b$ 
unique samples in the product vanishes. 
Hence,

\[ \mathbb{E} \Vert \nabla_{\theta} \bar{J}(\theta,\bar{x}) - \nabla \mathcal{J}(\theta) \Vert^2  =  \frac{1}{b} \sum_{s=1}^{b} \mathbb{E} \Vert  (\nabla_{\theta} J(\theta,x_s) - \nabla \mathcal{J}(\theta) )  \Vert^2 \leq \frac{\sigma^2}{b}.\]
Let $\bar{x}_j$ denote the $j^{th}$ mini-batch sampled from $\mathcal{R}^b$, i.e.,  $\bar{x}_j = ( x_{(j-1)b+1},\ldots,x_{jb} )$. 
It is clear from above that $\nabla_{\theta} \bar{J}(\theta,\bar{x}_j)$ has $\frac{\sigma^2}{b}$ bounded variance. Assuming that the update equation in line~\ref{upd} of Algorithm~\ref{algo1} has a regret bound of $\psi(\sigma^2,m)$ over $m$ entries of replay buffer, then the regret bound of $\bar{J}$ is given as
\[ \mathbb{E} [ \sum_{j=1}^{m/b} (\bar{J}(\theta_j,\bar{x}_j) - \bar{J}(\theta^*,\bar{x}_j))]   \leq \psi(\frac{\sigma^2}{b},\frac{m}{b}).\]
Using Equation~\ref{th3_1} and multiplying by $b$, we get
\begin{align*}
  \mathbb{E} [ \sum_{j=1}^{m/b} \sum_{i=(j-1)b+1}^{jb} (J(\theta_j, x_i) - J(\theta^*,x_i))]   \leq b\cdot \psi(\frac{\sigma^2}{b},\frac{m}{b}).
\end{align*}
Using $\psi(\sigma^2,m)=2D^2L+2D\sigma\sqrt{m}$, the regret bound in case of FAC is 
\[ 2bD^2L + 2D\sigma \sqrt{m}.\]
This shows that the leading term 
of regret bound of FAC is $\frac{b+\zeta^2+\zeta}{b}$ times lower than that of GAC algorithms which proves faster convergence via FAC algorithm.


\end{proof}
}

As mentioned earlier, a good mini-batch of data samples is one which has  independent and identically distributed data samples. In statistics, this means that the samples should be as random as possible. One way of quantifying randomness is using entropy~\cite{Shannon48} in data. In the following theorem, we prove that the FAC algorithm improves the IID-ness of the samples stored in the replay buffer using the notion of entropy.

\begin{theorem}
\label{thm3}
[Improving the IID characteristic of replay buffer]
    Let the replay buffer at a given instant consist of samples $( x_1,\ldots,x_m )$  with $\lambda$ equivalent elements for a GAC algorithm.
    Then the increase in the entropy of the replay buffer samples populated via the FAC algorithm with respect to the GAC algorithm  is given by
     \begin{align*}
        \Delta\mathcal{H}= \frac{(\lambda+1)  \cdot\texttt{log}\,(\lambda+1)}{m}.
    \end{align*}
\end{theorem}

\shortversion{
\begin{proof}
    Refer~\cite{singh24}.
\end{proof}
}
\longversion{
\begin{proof}
 Let $X = ( x_1,x_2,\ldots,x_m )$ be the batch samples used in line~\ref{upd} 
 of Algorithm~\ref{algo1}. Assume that replay buffer consists of more than two elements, i.e., $m>2$. Consider that this set contains $\lambda$ equivalent sample of $x_1$ such that $x_1=x_2=\ldots=x_{\lambda+1}$. In this proof, we have considered multiple duplicacy of a single element for the sake of brevity. In general, the replay buffer can have multiple duplicacy of multiple elements. We slightly abuse the notation $X$ to also represent it as a random variable to capture the distribution of replay buffer samples. Thus, the probability mass function of $X$ is given as:
\begin{equation}
  \mathbb{P}(X=x_i) =
    \begin{cases}
     \frac{1}{m} &~~~ \text{if $i \in [\lambda+2,m]$},\\
      \frac{\lambda+1}{m} &~~~ \text{otherwise}.\\
    \end{cases}       
\end{equation}

 The replay buffer populated via FAC contains only unique samples and hence  $X'=( x_{1},\ldots,x_{m})$. 
The probability mass function of $X'$ is given as:
   \begin{equation}
   \mathbb{P}(X'=x_i) = \frac{1}{m} ~~~\forall i \in [1,m].
      \end{equation}

 The entropy of $X$ is given as 
 \begin{align*}
     \mathcal{H}(X) &= -\sum_{i=1}^{m-\lambda} [\mathbb{P}(X_i)\cdot \texttt{log } \mathbb{P}(X_i)]\\
     &= - \left[\frac{\lambda+1}{m}\cdot \texttt{log }\frac{\lambda+1}{m} + \frac{1}{m}\cdot \texttt{log }\frac{1}{m} +\ldots+\frac{1}{m}\cdot \texttt{log }\frac{1}{m}  \right]\\
     &= - \left[\frac{\lambda+1}{m}\cdot \texttt{log }\frac{\lambda+1}{m} + \frac{m-\lambda-1}{m}\cdot \texttt{log }\frac{1}{m}\right]\\
     &= - \left[\frac{\lambda+1}{m}\cdot \texttt{log }\frac{\lambda+1}{m} - \frac{m-\lambda-1}{m}\cdot \texttt{log }m\right]\\
     &= -\frac{1}{m} [(\lambda+1)\cdot \texttt{log }(\lambda+1) - (\lambda+1)\texttt{log }m  - m\cdot \texttt{log }m  \\ 
     & \quad \quad   + (\lambda +1) \cdot \texttt{log }m ] \\
     &= \texttt{log }m -\frac{(\lambda+1)\cdot \texttt{log }(\lambda+1)}{m}. \numberthis
 \end{align*}


  The entropy of $X'$ is given as 
 \begin{align*}
     \mathcal{H}(X')  &= -\sum_{i=1}^{m} [\mathbb{P}(X_i)\cdot \texttt{log } \mathbb{P}(X_i)]\\ 
     &= - \left[\frac{1}{m}\cdot \texttt{log }\frac{1}{m} +\ldots+\frac{1}{m}\cdot \texttt{log }\frac{1}{m}  \right]\\ 
     &= - \left[\frac{m}{m}\cdot \texttt{log }\frac{1}{m} \right]\\
      &=  \texttt{log } (m). \\
 \end{align*}

 The random variable $X$ represents the replay buffer populated via GAC algorithm and $X'$ represents the replay buffer populated via FAC algorithm. Hence, in case replay buffer contains equivalent elements, i.e. $\zeta>0$,  we expect higher entropy for $X'$ than that of $X$. The difference in entropy is given as
 \begin{align*}
     \Delta \mathcal{H} &= \mathcal{H}(X') -\mathcal{H}(X)\\
     &= \texttt{log } (m) -\left( \texttt{log }m -\frac{(\lambda+1)\cdot \texttt{log }(\lambda+1)}{m}\right)\\
     &= \frac{(\lambda+1)\cdot \texttt{log }(\lambda+1)}{m}\\
    &>0. \numberthis
    \label{thm4_d}
 \end{align*}
\end{proof}
}


In Theorem~\ref{thm2}, we define convergence using a mini-batch of samples (size $b$) that is sampled from the replay buffer at any time step, whereas in Theorem~\ref{thm3}, we define IID characteristics over the whole replay buffer (size $m$). 

\section{Evaluation}

This section presents our experiments on synthesizing control policy using our proposed FAC algorithm. 

\subsection{Experimental Setup}
We implement our FAC algorithm as a generic algorithm that can be used on top of any off-policy actor-critic algorithm.
To simulate the environment, we use the $\texttt{gym}$ simulator~\cite{gym}.
We run all the synthesized control policies on 100 random seeds to measure the reward accumulation. 
All experiments are carried out on an Ubuntu22.04 machine with Intel(R) Xeon(R) Gold 6226R $@ 2.90$ GHz$\times16$ CPU, NVIDIA RTX A4000 32GB Graphics card, and 48\,GB RAM.
The source code of our implementation is available at \url{https://github.com/iitkcpslab/FAC}.

\subsubsection{Baselines.}
We use two state-of-the-art deep RL algorithms, SAC~\cite{haarnoja18} and TD3~\cite{fujimoto18}, developed by introducing specific enhancements to the GAC algorithm,  as the baseline for comparison with our proposed FAC algorithm as they are the most widely used model-free off-policy actor-critic algorithms for synthesizing controllers for “continuous” environments. While SAC uses stochastic policy gradient, TD3 uses deterministic policy gradient. Additionally, we compare FAC with state-of-the-art prioritization based algorithm LABER~\cite{lahire22}.

\subsubsection{Benchmarks.}
We use $9$ benchmarks based on Gym (Classic Control, Box2d, and Mujoco) in our experiments. The details of the benchmarks are provided in Table~\ref{table-bench}.

\begin{table}[t]
\begin{center}
\small
\begin{tabular} {lrrr}
\toprule
   Environment & \#Observations & \#Actions & $\kappa$ \\
  \midrule
  Pendulum & 3 & 1 & 1\\
  MCC & 2 & 1 & 1\\
  LLC & 8 & 2 & 2\\
  Swimmer  & 8 & 2 & 4\\
  Reacher & 11 & 2 & 5\\
  Hopper  & 11 & 3 & 4\\
   Walker & 17 & 6 & 5\\
  Ant & 27 & 8 & 8\\
  Humanoid & 376 & 17 & 16\\
 \bottomrule
\end{tabular}
\caption{Details of the Benchmarks}
\label{table-bench}
\end{center}
\end{table}



\begin{table*}
  \centering
  \begin{tabular}{l|rrr|rrr|rrr|r}
    \toprule
    Model    & $CP_{B}$ & $CP_{F}$ & $\Delta^{CP}$  &  $|\mathcal{R}_B|$ & $|\mathcal{R}_F|$ & $\Delta^{\mathcal{R}}$  & $R_B$ & $R_F$ & $\Delta^R$  & $P$\\
    \midrule
      Pendulum (SAC)  & $17600$ & $17600$ &  $-$ & $20000$ & $11412$ & $42.94$ & $-143.97_{\pm87.8
      0}$ & $-144.66_{\pm88.53}$ & $-0.47$  & $1.75$ \\
     Pendulum (TD3)  & $18400$ & $18400$ &  $-$ & $20000$ & $14137$ & $29.31$ & $-138.80_{\pm88.56}$ & $-136.89_{\pm85.41}$ & $+1.37$ & $1.44$ \\
     \midrule
         MC (SAC)  & $30696$  & $23089$ & $24.78$ & $50000$ & $10001$ &    $80.00$  & $-7.01_{\pm0.09}$  & $-7.09_{\pm0.41}$ & $-1.14$ & $4.94$  \\
     MC (TD3)  &  $38419$  & $33590$ & $12.56$  & $100000$ & $10004$ &  $90.00$ & $-6.13_{\pm0.22}$  & $-5.97_{\pm0.07}$ & $+2.61$  & $10.27$\\
\midrule
     LLC (SAC) &  $315118$ & $215108$  & $31.73$ & $500000$ & $422642$ & $15.47$ & $182.28_{\pm17.50}$ & $179.72_{\pm16.05}$ & $-1.40$  & $1.17$ \\
     LLC (TD3) &  $290846$ & $263927$  & $9.25$  & $200000$ & $70026$ & $65.00$  & $137.90_{\pm51.81}$ & $146.43_{\pm26.90}$ & $+6.18$  & $3.03$\\
\midrule
Swimmer (SAC) & $232000$  & $188000$ & $18.96$  & $500000$ & $83922$ &    $83.21$  & $338.11_{\pm1.95}$  & $328.11_{\pm1.98}$ & $-2.95$  & $5.82$ \\
     Swimmer (TD3)   &  $216000$  & $928000$ & $-$   & $1000000$ & $61289$ &    $93.87$  & $47.45_{\pm1.37}$  & $114.86_{\pm1.91}$ & $+142.06 $ & $46.75$ \\
\midrule
     Reacher (SAC)  &  $183600$  & $171600$ & $6.53$ & $300000$  & $151743$ & $49.41$  & $19.18_{\pm10.70}$ & $19.96_{\pm10.23}$  & $+4.06$ & $2.17$ \\
     Reacher (TD3) & $249600$  & $243600$ & $2.40$  & $20000$ & $20000$ & $0$  & $13.45_{\pm12.10}$ & $17.10_{\pm10.28}$  & $+27.13$  & $1.40$  \\
\midrule
    Hopper (SAC) &  $725191$ & $577319$  & $20.39$  & $1000000$ & $290327$ & $70.96$ & $3273.10_{\pm5.98}$ & $3529.13_{\pm6.49}$  & $+7.82$  & $3.72$ \\
    Hopper (TD3)  & $998851$ & $848361$  & $15.06$  & $1000000$ & $320182$ & $67.98$ & $3232.16_{\pm246.55}$ & $3269.26_{\pm562.92}$  & $+1.14$ & $3.16$ \\
\midrule
    Walker (SAC) & $986412$  & $779446$ & $20.98$  & $1000000$ & $696091$ &   $30.39$   & $4332.53_{\pm546.30}$  & $5386.14_{\pm24.75}$ & $+24.33$  & $1.79$\\
     Walker (TD3)  & $784014$  & $855112$ & $-$  & $1000000$ & $755460$ &   $24.45$  & $4517.75_{\pm9.86}$  & $4875.14_{\pm57.47}$ & $+7.91$ & $1.43$ \\
\midrule
    Ant (SAC)  & $909250$  & $782363$ & $13.95$   & $1000000$ & $872183$ &   $12.78$  & $3800.34_{\pm1289.42}$  & $5057.42_{\pm1022.72}$ & $+33.08$ & $1.52$   \\
      Ant (TD3)  &   $784967$  & $676999$ & $13.75$    & $1000000$ & $754430$ &   $24.55$  & $5397.40_{\pm400.47}$  & $5416.09_{\pm139.93}$ & $+0.34$  &  $1.33$  \\
\midrule
     Humanoid (SAC) & $1819768$  & $1079479$ & $40.68$   & $1000000$  & $807923$ &    $19.20$  & $5706.61_{\pm639.32}$  & $5479.23_{\pm786.85}$ & $-3.98$  & $1.19$\\
      Humanoid (TD3)  & $1841538$  & $1320449$ & $28.29$   & $1000000$  & $436749$ &    $56.32$ & $5231.64_{\pm695.39}$  & $5217.76_{\pm396.35}$ & $-0.26 $ & $2.29$ \\
    \bottomrule
  \end{tabular}
    \caption{Comparison of Convergence (CP), Replay buffer size ($\mathcal{R}$) and Reward (R) of FAC with baselines SAC and TD3}
  \label{table-full}
\end{table*}

\subsection{Performance Metrics}

Here, we present the performance metrics we use to evaluate FAC in comparison to the baseline algorithms.

\subsubsection{Convergence.} 
We define convergence point (CP) as the timestep at which reward improvement settles down, i.e., it enters a band of reward range $[0.9\cdot r_{max},r_{max}]$ and remains within it thereafter till the end of learning. Here, $r_{max}$ refers to the maximum reward attained during the training.

We define the percentage improvement in convergence ($\Delta^{CP}$) achieved by FAC ($CP_F$) w.r.t. baselines ($CP_B$)  as: 
\begin{equation}
    \Delta^{CP} = \frac{CP_{B}-CP_{F}}{CP_{B}}\times 100.
\end{equation}

\subsubsection{Size of Replay Buffer.}
The size of a replay buffer $|\mathcal{R}|$ is the number of transitions stored in it at the completion of learning.
The replay buffer in the baseline setting 
is denoted by $\mathcal{R}_B$ and w.r.t the FAC Algorithm by $\mathcal{R}_F$.
We measure the reduction in size of replay buffer $\Delta^\mathcal{R}$ achieved via FAC w.r.t. a baseline as:
\begin{equation}
    \Delta^{\mathcal{R}} = \frac{|\mathcal{R}_B|- |\mathcal{R}_F|}{|\mathcal{R}_{B}|}\times 100.
\end{equation}

\subsubsection{Reward.} 
The total reward accumulated by the synthesized control policy in the baseline setting 
is denoted by $R_B$ and for the FAC Algorithm by $R_F$.
The percentage change in  total reward accumulated $\Delta^R$ by FAC algorithm  w.r.t. baseline algorithm is defined as
\begin{equation}
    \Delta^R = \frac{R_{F}- R_{B}}{R_{B}}\times 100.
\end{equation}

\subsubsection{Sample Efficiency.}
We define this metric to capture the combined effect of reward improvement and the reduction in the size of the replay buffer.
We denote the per-sample efficiency for FAC using $P_F$ and that of baseline using $P_B$ as \mbox{$P_F = R_F/|\mathcal{R}_F|$} and  $P_B = R_B/|\mathcal{R}_B|$.

Now we define a metric $P$ to capture the relative per-sample efficiency achieved by FAC w.r.t baseline algorithm as
\begin{equation}
    P = \frac{P_F}{P_B} = 
   \frac{R_F\cdot |\mathcal{R}_B|}{|\mathcal{R}_F|\cdot R_B}.
\end{equation}
  
Note that the above equation is meaningful if the rewards are positive. In the case of negative rewards, we first translate them into positive values by adding to them the sum of the absolute values of the reward in both cases. For instance, in the case of pendulum w.r.t. SAC (see Table~\ref{table-full}), the addition term is ${\tt abs}(-143.97)+{\tt abs}(-144.66) = 288.63$. So, the translated reward for baseline ($R_B$) is $144.66$, and for FAC is $143.97$.

\smallskip
\noindent
\subsection{Hyper-parameters}
We run the experiments using the same configuration as used in the baselines~\cite{stable-baselines3}, i.e., the state-of-the-art GAC algorithms SAC and TD3.
In Equation~\ref{kde}, we use the Epanechnikov kernel~\cite{jones90} function. 
The values of other hyper-parameters used in experiments are shown in Table~\ref{table-hp}.

\begin{table}[t]
\begin{center}
\begin{tabular} {ccccc}
\toprule
   $\nu$ & $\epsilon$ & $\eta$ & $\beta$  & $\mu$  \\
  \midrule
  $0.5$ & $0.2$ & $1e5$   & $0.2$  & $50$\\
 \bottomrule
\end{tabular}
\caption{Hyper-parameters used in the experiments}
\label{table-hp}
\end{center}
\end{table}





We consider three kernels for our density estimation, namely Gaussian, Tophat, and Epanechnikov.
In general, the curve corresponding to the Gaussian kernel is too smooth, while that of the Tophat kernel is quite coarse. Thus, we select the Epanechnikov kernel as the curve is well balanced w.r.t. smoothness and coarseness.
We perform experiments on three benchmark systems to judge the efficacy of the kernels.
\shortversion{
We observe that the Epanechnikov kernel indeed provides the best results consistently among  these three kernels in terms of the accumulated rewards (refer~\cite{singh24} for the details).
}
\longversion{
We observe that the Epanechnikov kernel indeed provides the best results consistently among all these three kernels in terms of the accumulated rewards. We show the training plots for different kernel functions in Figure~\ref{fig:kcomp}.
\begin{figure}[!htbp]
      \centering
     \begin{subfigure}[b]{0.156\textwidth}
        \centering
         \includegraphics[width=\textwidth]{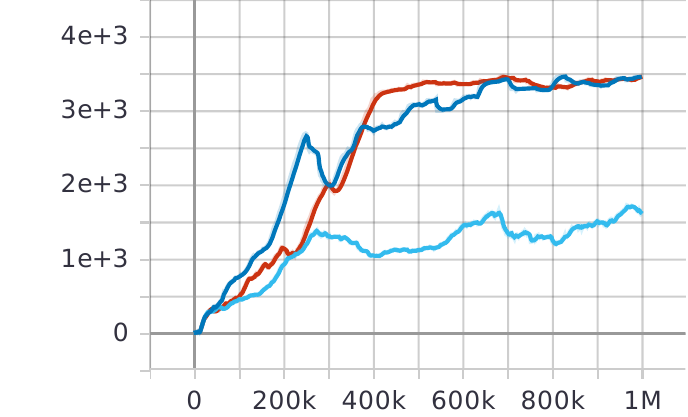}
         \caption{\textsf{Hopper}}
         \label{fig:hk}
     \end{subfigure}
     \begin{subfigure}[b]{0.156\textwidth}
         \centering
         \includegraphics[width=\textwidth]{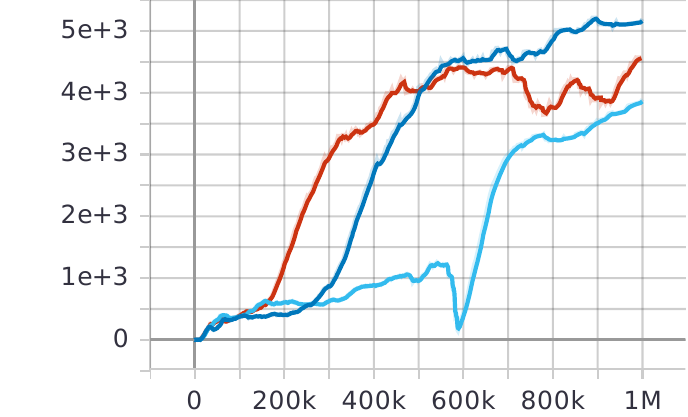}
         \caption{\textsf{Walker}}
         \label{fig:wk}
    \end{subfigure}
     \begin{subfigure}[b]{0.156\textwidth}
         \centering
         \includegraphics[width=\textwidth]{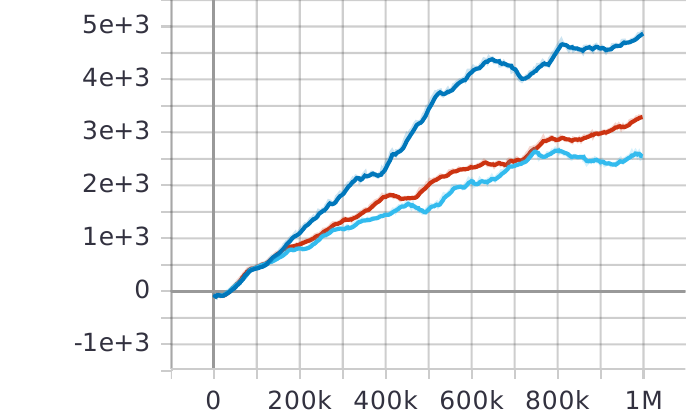}
         \caption{\textsf{Ant}}
         \label{fig:ak}
    \end{subfigure}
    \caption{Plots showing comparison of epanechnikov (dark blue), tophat (dark red) and gaussian (sky blue) kernels.}
    \label{fig:kcomp}   
\end{figure}
}

\shortversion{
The results of experiments for deciding the other hyper-parameters values are available in~\cite{singh24}.
}
\longversion{

In Table~\ref{table-hp}, the value of $\nu=0.5$ indicates that we classify those dimensions as important for which the corresponding singular value is at least $50\%$ of the largest singular value. This is to maintain a balance between removing redundant dimensions and retaining important dimensions.
We show the training plots for different values of hyper-parameters $\epsilon$, $\eta$, $\beta$ and $\mu$ in Figures~\ref{fig:epcomp}-\ref{fig:mcomp}. Note that in all the plots, the plot corresponding to the best value of a hyper-parameter is shown in \textit{dark blue}.

In the case of $\epsilon$, we observe that control policy synthesis is worst for value $0.1$. This is because, for this value, it becomes too conservative and starts throwing away even the important samples. For large values of $\epsilon$, the synthesis is sub-optimal due to the presence of large numbers of equivalent samples (as overall lesser samples are discarded).

\begin{figure}[!htbp]
      \centering
     \begin{subfigure}[b]{0.153\textwidth}
        \centering
         \includegraphics[width=\textwidth]{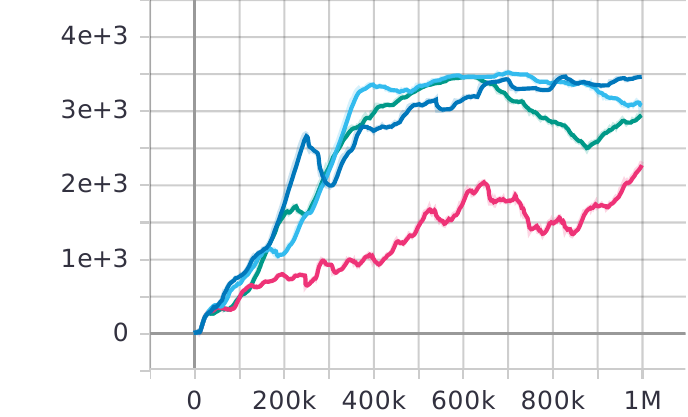}
         \caption{\textsf{Hopper}}
         \label{fig:he}
     \end{subfigure}
     \begin{subfigure}[b]{0.153\textwidth}
         \centering
         \includegraphics[width=\textwidth]{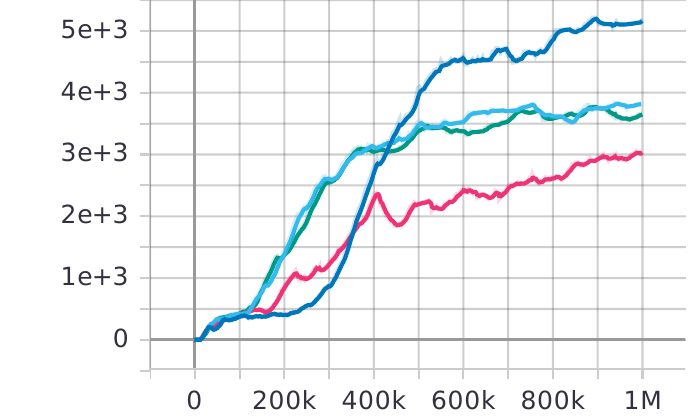}
         \caption{\textsf{Walker}}
         \label{fig:we}
    \end{subfigure}
     \begin{subfigure}[b]{0.153\textwidth}
         \centering
         \includegraphics[width=\textwidth]{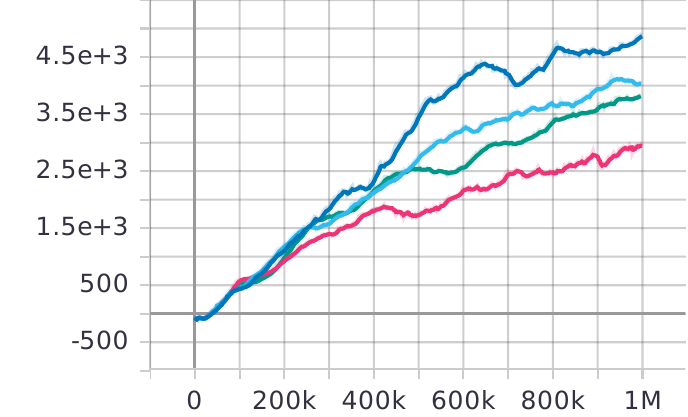}
         \caption{\textsf{Ant}}
         \label{fig:ae}
    \end{subfigure}
    \caption{Plots showing training for different values of $\epsilon$ - 0.1 (pink), 0.2 (dark blue), 0.3 (green), and  0.5 (light blue).}
    \label{fig:epcomp}   
\end{figure}

In the case of $\eta$, for the least value $10k$ (light blue), the synthesized control policy is among the worst. An interesting trend is witnessed for value $50k$ (pink), which performs poorly for Hopper but is close to best for Walker. This is due to the fact that the maximum number of distinct rewards for an abstract state is around $70k$ in Hopper and 
$13k$ in Walker.

\begin{figure}[!htbp]
      \centering
     \begin{subfigure}[b]{0.153\textwidth}
        \centering
         \includegraphics[width=\textwidth]{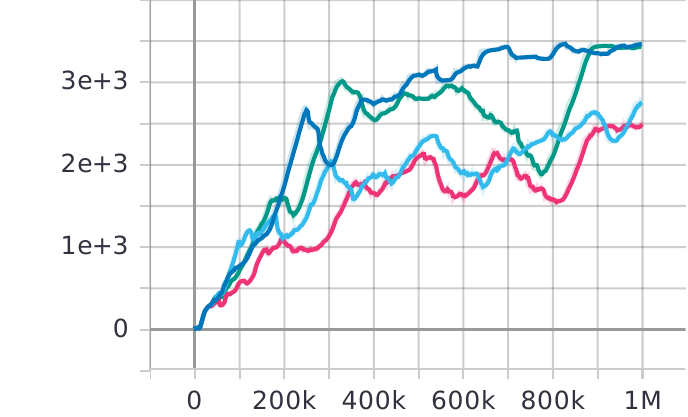}
         \caption{\textsf{Hopper}}
         \label{fig:het}
     \end{subfigure}
     \begin{subfigure}[b]{0.153\textwidth}
         \centering
         \includegraphics[width=\textwidth]{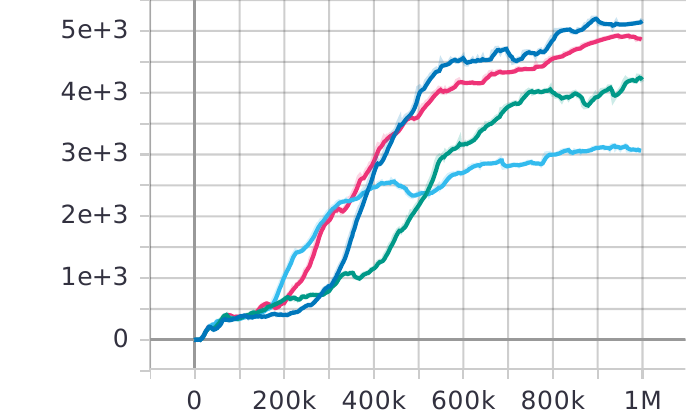}
         \caption{\textsf{Walker}}
         \label{fig:wet}
    \end{subfigure}
     \begin{subfigure}[b]{0.153\textwidth}
         \centering
         \includegraphics[width=\textwidth]{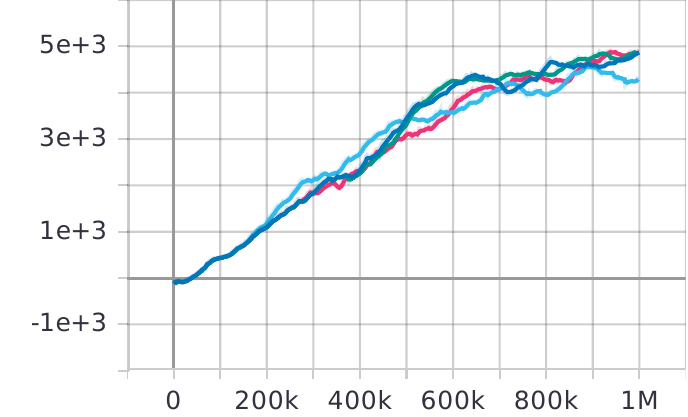}
         \caption{\textsf{Ant}}
         \label{fig:aet}
    \end{subfigure}
    \caption{Plots showing training for different values of $\eta$ - 10k (light blue), 50k (pink), 100k (dark blue), and 150k (green).}
    \label{fig:etacomp}   
\end{figure}

The parameter $\beta$ is like the radius of a region whose density is calculated to check if a new sample can be added or not.
For value $0.5$ (light blue), the synthesized control policy is among the worst. This is because we check the density of a large region, and quite possibly, it would be higher (hence discard samples) even though it could accommodate more unique samples. For value $0.1$ (green), the density is estimated for a very small region. This leads to the accommodation of more samples than needed, i.e., a closer equivalent sample can be added as it doesn't lie inside the region.

\begin{figure}[!htbp]
      \centering
     \begin{subfigure}[b]{0.153\textwidth}
        \centering
         \includegraphics[width=\textwidth]{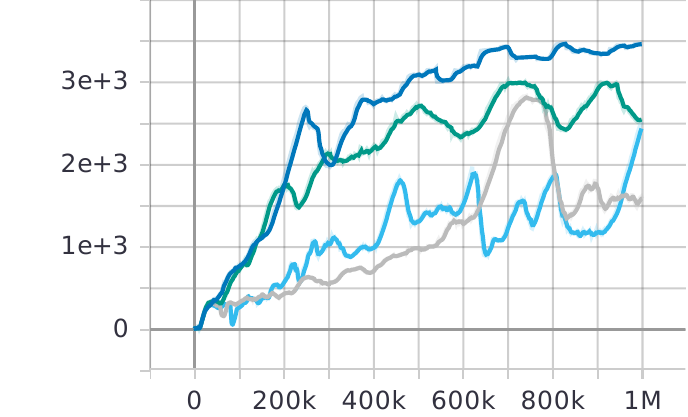}
         \caption{\textsf{Hopper}}
         \label{fig:hb}
     \end{subfigure}
     \begin{subfigure}[b]{0.153\textwidth}
         \centering
         \includegraphics[width=\textwidth]{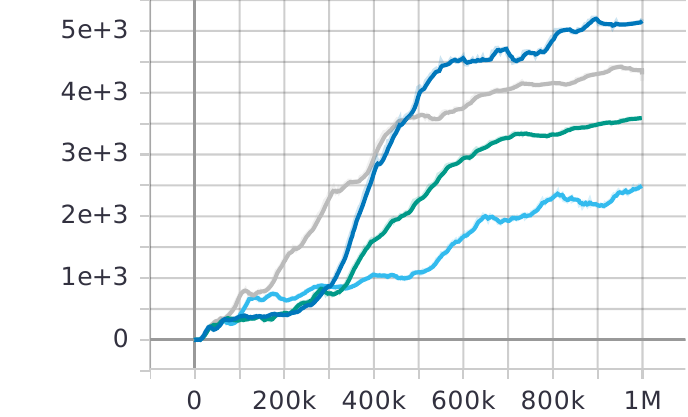}
         \caption{\textsf{Walker}}
         \label{fig:wb}
    \end{subfigure}
     \begin{subfigure}[b]{0.153\textwidth}
         \centering
         \includegraphics[width=\textwidth]{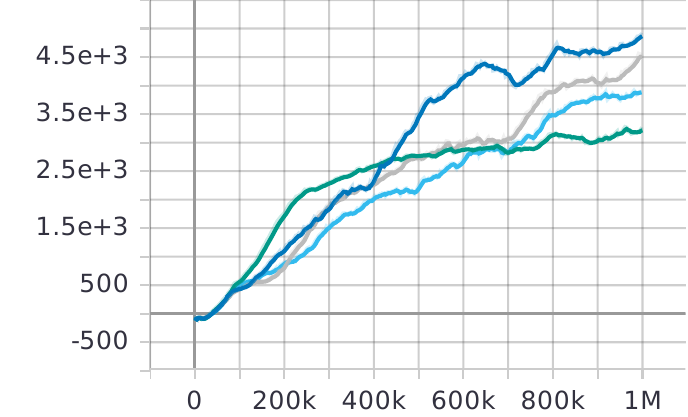}
         \caption{\textsf{Ant}}
         \label{fig:ab}
    \end{subfigure}
    \caption{Plots showing training for different values of $\beta$ - 0.1 (green), 0.2 (dark blue), 0.3 (gray), and 0.5 (light blue).}
    \label{fig:bcomp}   
\end{figure}

For $\mu$, the worst control policy is synthesized when the size of the abstract state is large (and hence less number of abstract states). This means discarding too many samples as they would be classified as equivalent. This is witnessed for $\mu=30$. The best control policy was synthesized for $\mu=50$, and thereafter, it gets saturated. This is because a smaller size of abstract states would lead to the classification of even equivalent states as distinct states.

\begin{figure}[!htbp]
      \centering
     \begin{subfigure}[b]{0.153\textwidth}
        \centering
         \includegraphics[width=\textwidth]{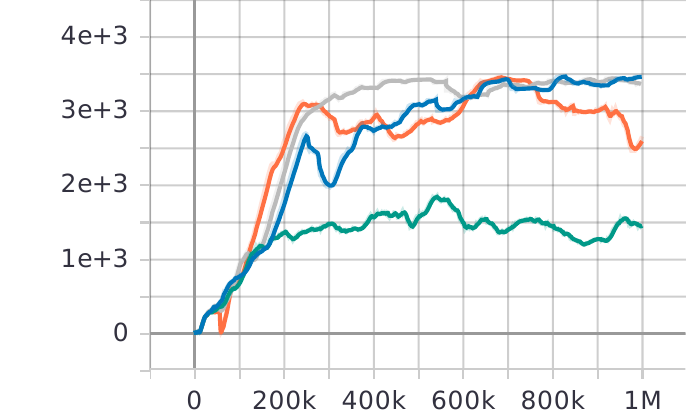}
         \caption{\textsf{Hopper}}
         \label{fig:hm}
     \end{subfigure}
     \begin{subfigure}[b]{0.153\textwidth}
         \centering
         \includegraphics[width=\textwidth]{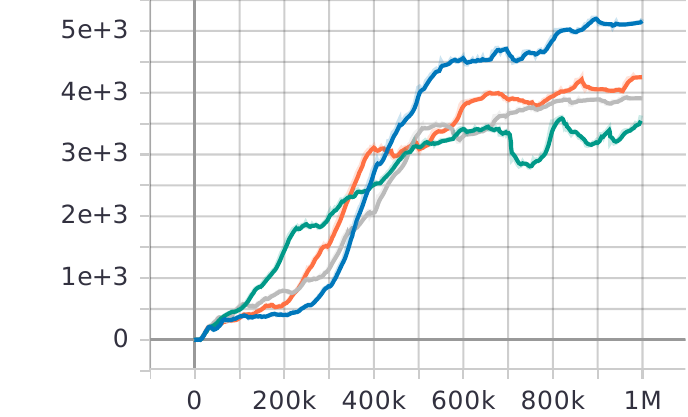}
         \caption{\textsf{Walker}}
         \label{fig:wm}
    \end{subfigure}
     \begin{subfigure}[b]{0.153\textwidth}
         \centering
         \includegraphics[width=\textwidth]{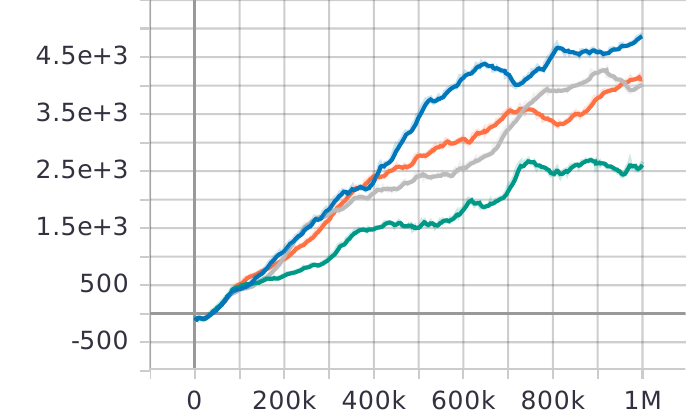}
         \caption{\textsf{Ant}}
         \label{fig:am}
    \end{subfigure}
    \caption{Plots showing training for different values of $\mu$ - 30 (green), 50 (dark blue), 70 (orange), and 100 (gray).}
    \label{fig:mcomp}   
\end{figure}

}

\subsection{Results}
In this subsection, we present our experimental results.

\smallskip
\noindent
\textbf{Convergence, Replay Buffer Size, and Accumulate Reward.}
\shortversion{
The training for control policy synthesis (refer to the plots in~\cite{singh24}) is done using the same configuration as used in the stable-baselines3~\cite{stable-baselines3}.
}
\longversion{
In Figure~\ref{fig:test3} and~\ref{fig:test4}, we present the comparison of FAC and LABER algorithm w.r.t SAC and TD3, respectively. Each plot shows the improvement in reward accumulated with each time step.  
\begin{figure}[!htbp]
      \centering
     \begin{subfigure}[b]{0.15\textwidth}
        \centering
         \includegraphics[width=\textwidth]{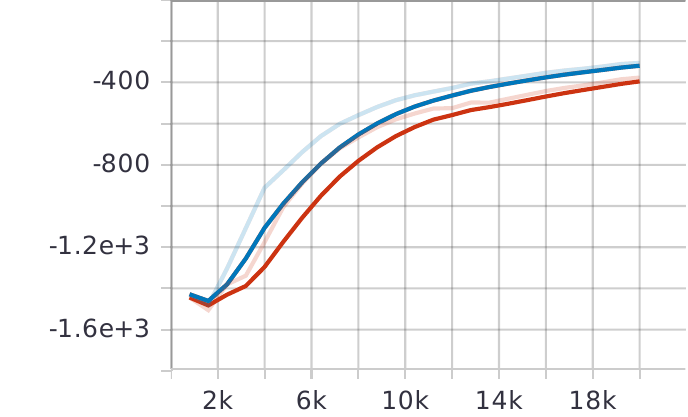}
         \caption{\textsf{Pendulum}}
         \label{fig:pend1}
     \end{subfigure}
     \begin{subfigure}[b]{0.15\textwidth}
         \centering
         \includegraphics[width=\textwidth]{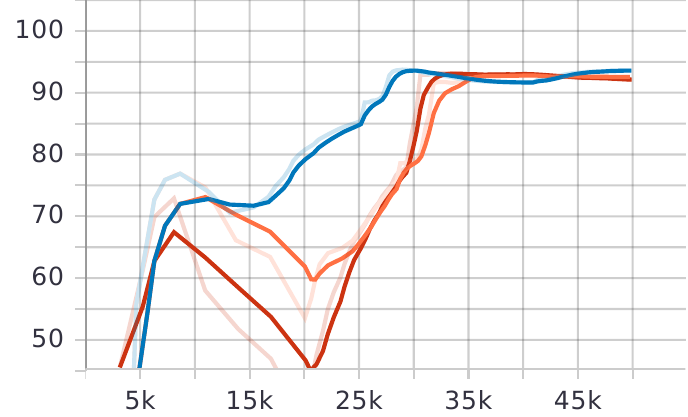}
         \caption{\textsf{MCar}}
         \label{fig:mc1}
    \end{subfigure}
     \begin{subfigure}[b]{0.15\textwidth}
         \centering
         \includegraphics[width=\textwidth]{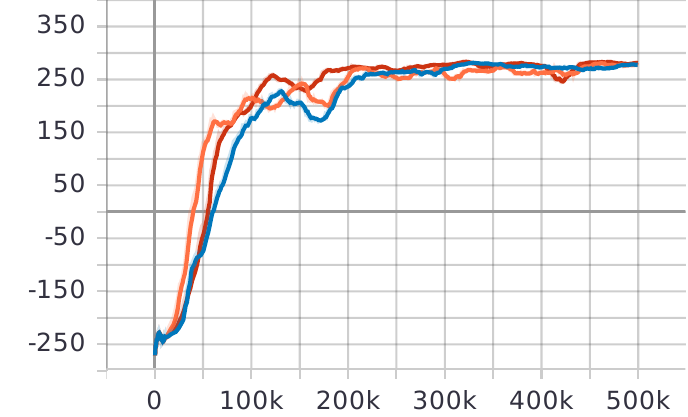}
         \caption{\textsf{LLC}}
         \label{fig:llc1}
    \end{subfigure}
    \begin{subfigure}[b]{0.15\textwidth}
        \centering
         \includegraphics[width=\textwidth]{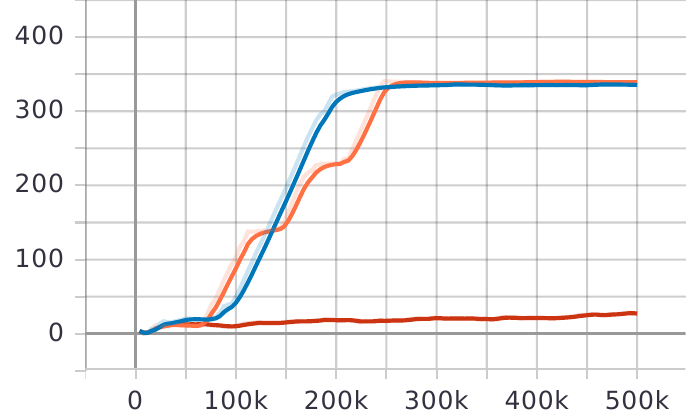}
         \caption{\textsf{Swimmer}}
         \label{fig:swm1}
     \end{subfigure}
     \begin{subfigure}[b]{0.15\textwidth}
         \centering
         \includegraphics[width=\textwidth]{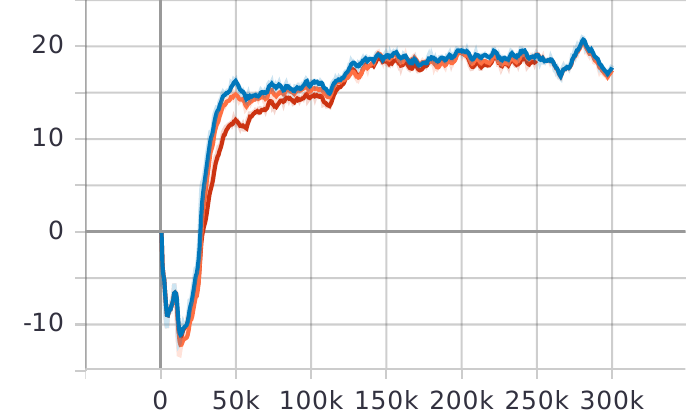}
         \caption{\textsf{Reacher}}
         \label{fig:rr1}
    \end{subfigure}
     \begin{subfigure}[b]{0.15\textwidth}
         \centering
         \includegraphics[width=\textwidth]{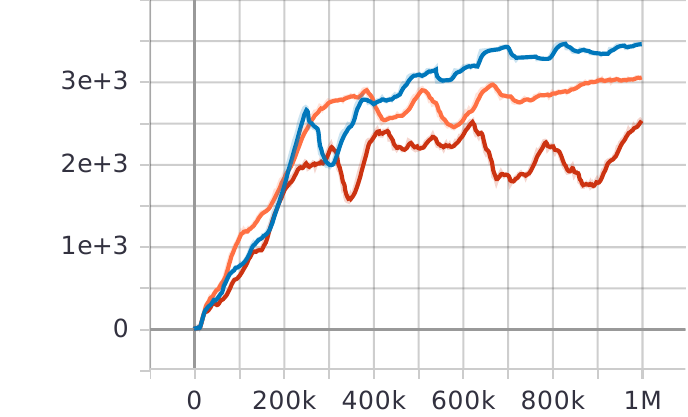}
         \caption{\textsf{Hopper}}
         \label{fig:hop1}
    \end{subfigure}
    \begin{subfigure}[b]{0.15\textwidth}
        \centering
         \includegraphics[width=\textwidth]{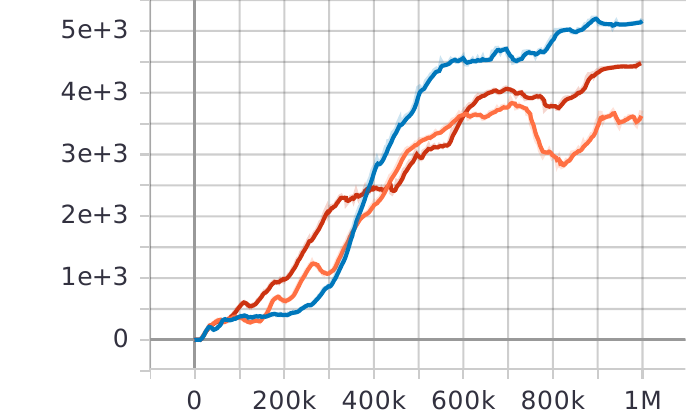}
         \caption{\textsf{Walker}}
         \label{fig:wkr1}
     \end{subfigure}
     \begin{subfigure}[b]{0.15\textwidth}
         \centering
         \includegraphics[width=\textwidth]{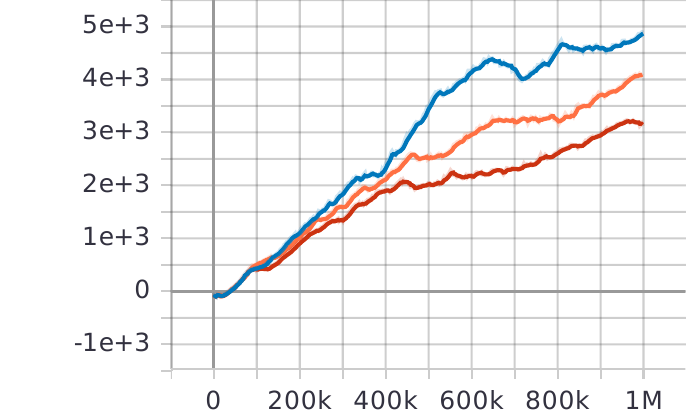}
         \caption{\textsf{Ant}}
         \label{fig:ant1}
    \end{subfigure}
     \begin{subfigure}[b]{0.15\textwidth}
         \centering
         \includegraphics[width=\textwidth]{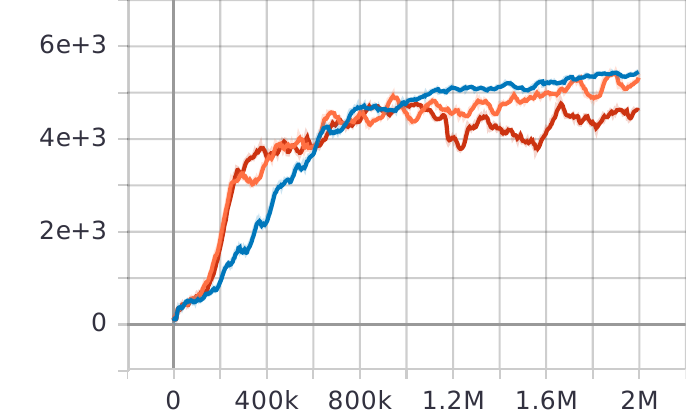}
         \caption{\textsf{Humanoid}}
         \label{fig:hum1}
    \end{subfigure}
     \caption{Training plots for SAC: baseline (orange) v/s LABER (red) v/s FAC (blue).}
    \label{fig:test3}
\end{figure}
\begin{figure}[!htbp]
      \centering
     \begin{subfigure}[b]{0.15\textwidth}
        \centering
         \includegraphics[width=\textwidth]{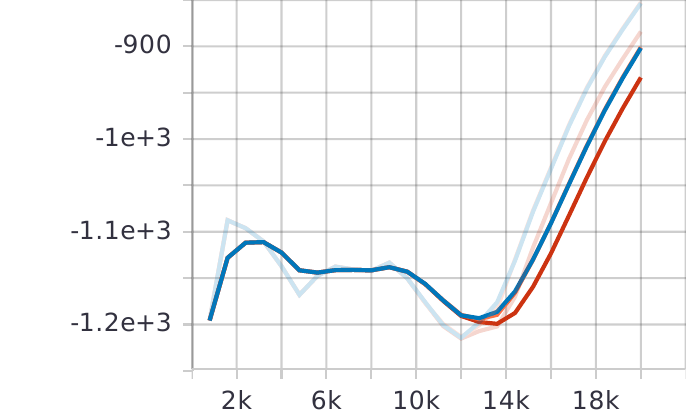}
         \caption{\textsf{Pendulum}}
         \label{fig:pend2}
     \end{subfigure}
     \begin{subfigure}[b]{0.15\textwidth}
         \centering
         \includegraphics[width=\textwidth]{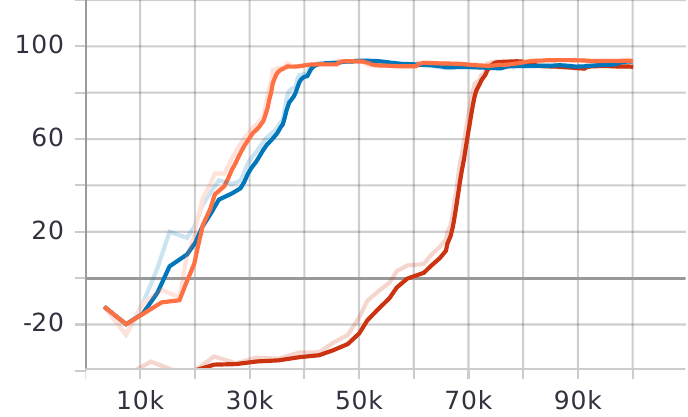}
         \caption{\textsf{MCar}}
         \label{fig:mc2}
    \end{subfigure}
     \begin{subfigure}[b]{0.15\textwidth}
         \centering
         \includegraphics[width=\textwidth]{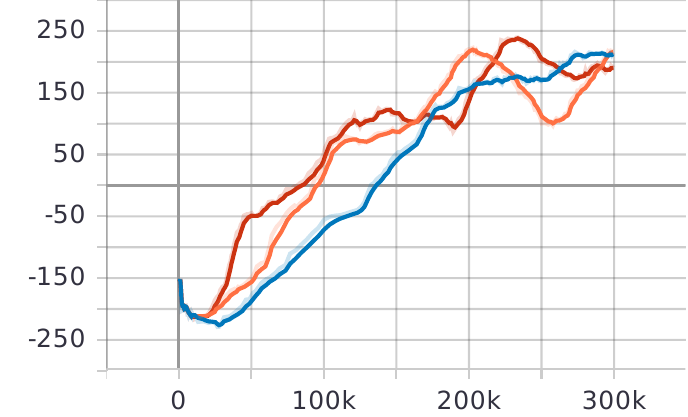}
         \caption{\textsf{LLC}}
         \label{fig:llc2}
    \end{subfigure}
    \begin{subfigure}[b]{0.15\textwidth}
        \centering
         \includegraphics[width=\textwidth]{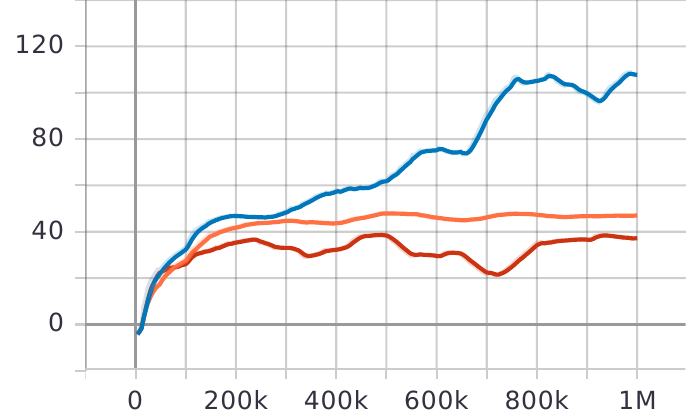}
         \caption{\textsf{Swimmer}}
         \label{fig:swm2}
     \end{subfigure}
     \begin{subfigure}[b]{0.15\textwidth}
         \centering
         \includegraphics[width=\textwidth]{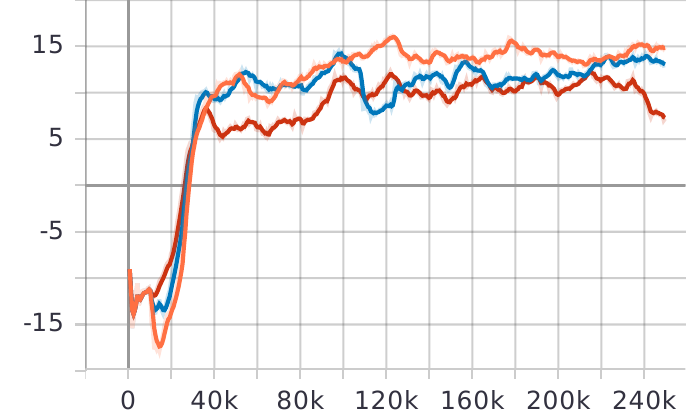}
         \caption{\textsf{Reacher}}
         \label{fig:rr2}
    \end{subfigure}
     \begin{subfigure}[b]{0.15\textwidth}
         \centering
         \includegraphics[width=\textwidth]{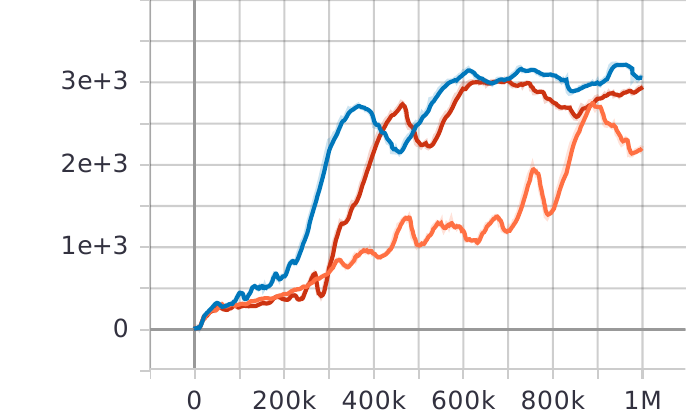}
         \caption{\textsf{Hopper}}
         \label{fig:hop2}
    \end{subfigure}
    \begin{subfigure}[b]{0.15\textwidth}
        \centering
         \includegraphics[width=\textwidth]{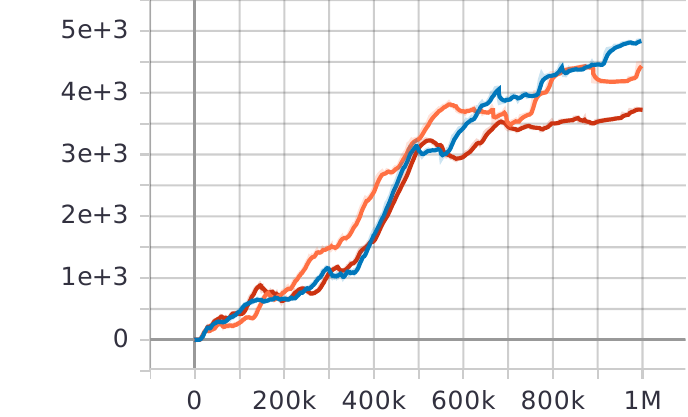}
         \caption{\textsf{Walker}}
         \label{fig:wkr2}
     \end{subfigure}
     \begin{subfigure}[b]{0.15\textwidth}
         \centering
         \includegraphics[width=\textwidth]{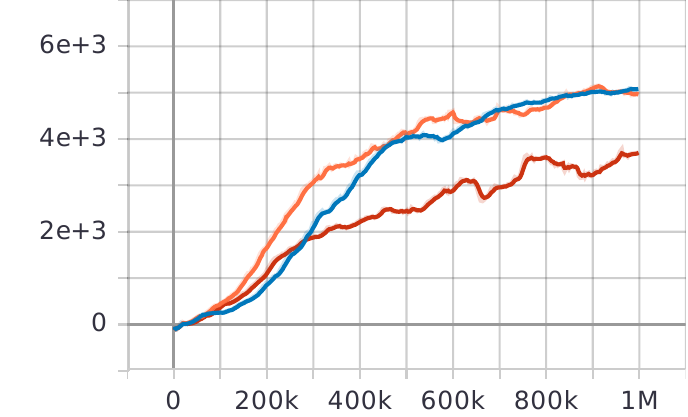}
         \caption{\textsf{Ant}}
         \label{fig:ant2}
    \end{subfigure}
     \begin{subfigure}[b]{0.15\textwidth}
         \centering
         \includegraphics[width=\textwidth]{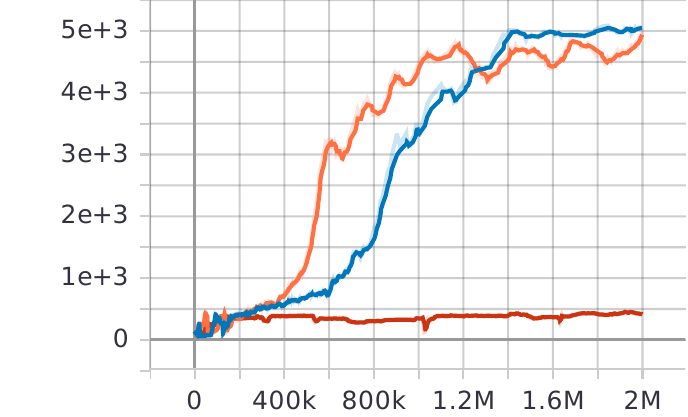}
         \caption{\textsf{Humanoid}}
         \label{fig:hum2}
    \end{subfigure}
     \caption{Training plots for TD3: baseline (orange) v/s LABER (red) v/s FAC (blue).}
    \label{fig:test4}
\end{figure}
}
Table~\ref{table-full} compares
the convergence, size of the replay buffer, total accumulated reward, and sample efficiency of FAC with those of SAC and TD3 on nine benchmarks. From the Table, we can see the following trends:
\begin{itemize}
    \item FAC converges faster than both SAC and TD3 on most of the benchmarks. In comparison to SAC and TD3, it converges up to $40\%$  and $28\%$ faster, respectively.
    \item FAC consistently requires a significantly smaller replay buffer compared to both SAC and TD3. It achieves up to $83\%$ reduction in the size of the replay buffer in comparison to SAC and up to $94\%$ reduction in comparison to TD3. For instance, the memory requirement of the replay buffer in the case of Humanoid is $\texttt{5.7GB}$ for TD3 and $\texttt{2.48GB}$ for FAC.
    \item For several benchmarks, FAC achieves significantly better accumulated reward. In the remaining cases, the degradation is marginal (up to only $4\%$).
    \item For all the benchmarks, FAC convincingly outperforms both SAC and TD3 in terms of sample efficiency.
\end{itemize}



However, there are a few exceptions to the general trends. For instance, 
for the simplest case of {\small \textsf{Pendulum}} in which the duration of control policy synthesis is small, we do not observe any improvement in convergence or the accumulated reward with FAC, but we observe a significant reduction in replay buffer size (upto $43\%$).
In the case of {\small \textsf{Swimmer}}, the FAC requires significantly more steps to converge than TD3. However,
no useful control policy could be synthesized for baseline TD3, as evidenced by low accumulated reward. On the other hand, the FAC extension of TD3 could synthesize a significantly superior controller with an improvement of $142\%$ in the accumulated reward and almost $94\%$ smaller replay buffer.
Similarly, in the case of {\small \textsf{Walker}} w.r.t. baseline TD3, convergence was slightly delayed on account of improvement in the size of the replay buffer and the reward accumulation. In the case of {\small \textsf{Reacher}}, w.r.t. baseline TD3, there was no reduction in replay buffer size as the baseline replay buffer size was already very small. However, we get $27\%$ improvement in reward accumulation and improved convergence.

\smallskip
\noindent
\textbf{Control Cost.}
In general, we observe a decrease in control cost for FAC w.r.t. the baselines (refer Table~\ref{table-cc}). In some cases, there has been an increase in the control cost (shown in red), but it led to the synthesis of a superior control policy with more reward accumulation (see Table~\ref{table-full}).   

\begin{table}
  \centering
  \begin{tabular}{lrr}
    \toprule
    Model    & Baseline & FAC   \\
    \midrule
    Pendulum (SAC) & $180.24\pm32.33$ & $72.49\pm42.16$  \\
     Pendulum (TD3) & $135.46\pm52.47$ & $119.08\pm48.45$  \\
      \midrule
    MCC(SAC)  & $76.06\pm17.51$  & $70.99\pm4.19$ \\
    MCC(TD3)  & $61.32\pm2.21$  & $59.78\pm0.79$ \\
    \midrule
    LLC (SAC) & $141.68\pm105.20$ & $113.64\pm59.35$  \\
    LLC (TD3) & $257.93\pm295.78$ & $182.72\pm110.07$  \\
    \midrule
        Swimmer(SAC)  & $1394.31\pm8.86$  & $1264.12\pm4.99$ \\
      Swimmer(TD3)  & $1135.26\pm224.99$  & {\color{red} $1340.31\pm22.94$} \\
     \midrule
    Reacher(SAC)  & $13.20\pm8.86$  & $13.30\pm8.88$ \\
    Reacher(TD3)  & $69.51\pm56.77$  & $40.18\pm32.89$ \\
    \midrule
         Hopper(SAC) & $969.93\pm11.94$ & {\color{red} $1025.52\pm12.55$}  \\
    Hopper(TD3) & $1390.49\pm119.07$ & $1325.62\pm221.50$  \\
    \midrule
    Walker(SAC)  & $2992.32\pm328.23$  & {\color{red} $3169.07\pm16.37$} \\
    Walker(TD3)  & $4901.46\pm14.49$  & $4592.26\pm476.37$ \\
    \midrule
    Ant(SAC)  & $2166.00\pm602.51$  & {\color{red} $2205.39\pm401.49$} \\
    Ant(TD3)  & $2327.36\pm156.66$  & $2186.27\pm42.31$ \\
    \midrule
     Humanoid(SAC)  & $1415.79\pm146.36$  & $1065.29\pm121.39$ \\
     Humanoid(TD3)  & $1953.06\pm246.69$  & $1807.19\pm131.33$ \\
    \bottomrule
  \end{tabular}
    \caption{Comparison of control cost of the policy synthesized via FAC w.r.t SAC and TD3}
  \label{table-cc}
\end{table}

\begin{figure}[!htbp]
      \centering
     \begin{subfigure}[b]{0.225\textwidth}
        \centering
         \includegraphics[width=\textwidth]{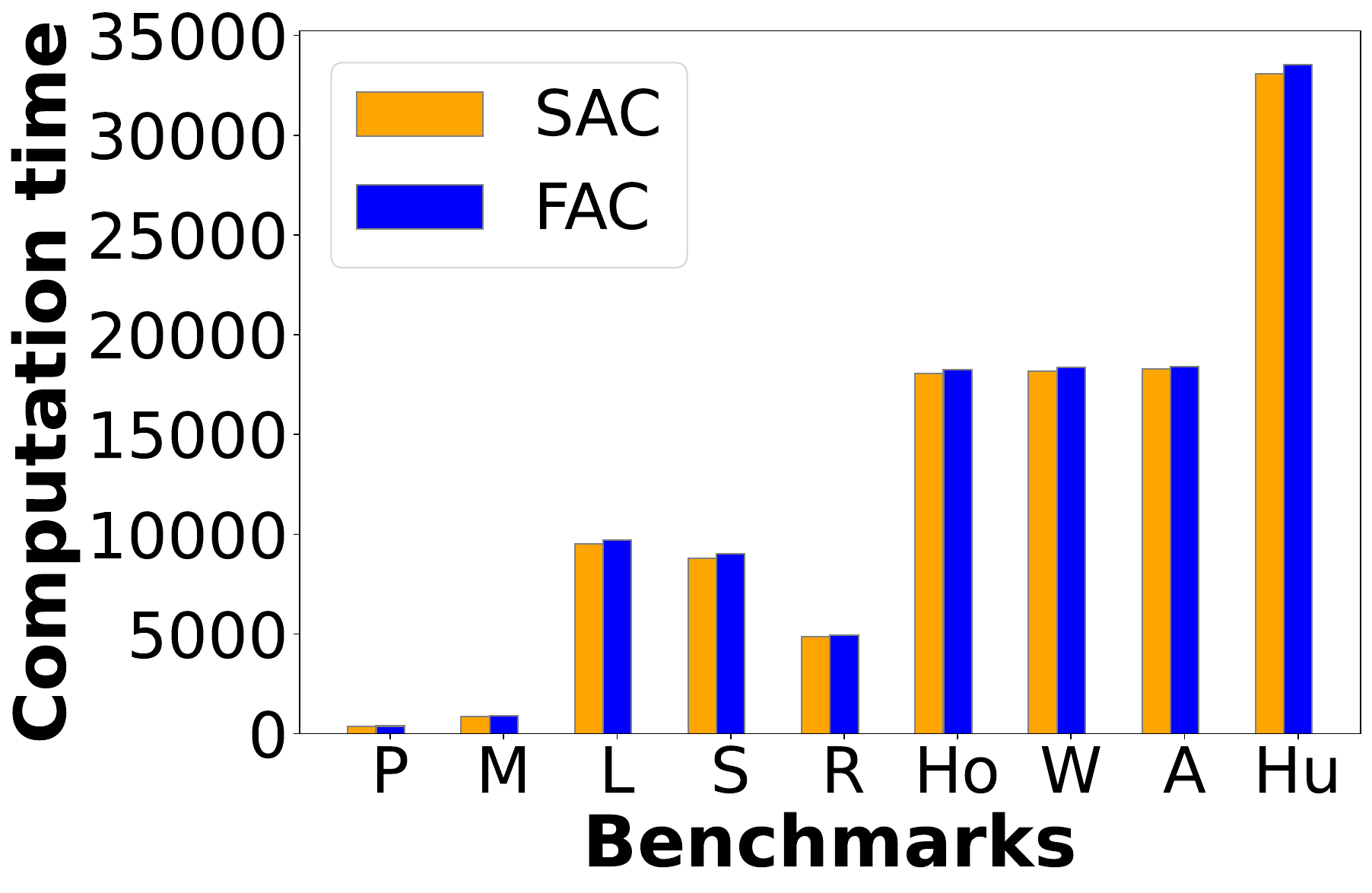}
         \caption{\textsf{FAC v/s SAC}}
         \label{fig:cts}
     \end{subfigure}
     \quad
     \begin{subfigure}[b]{0.225\textwidth}
         \centering
         \includegraphics[width=\textwidth]{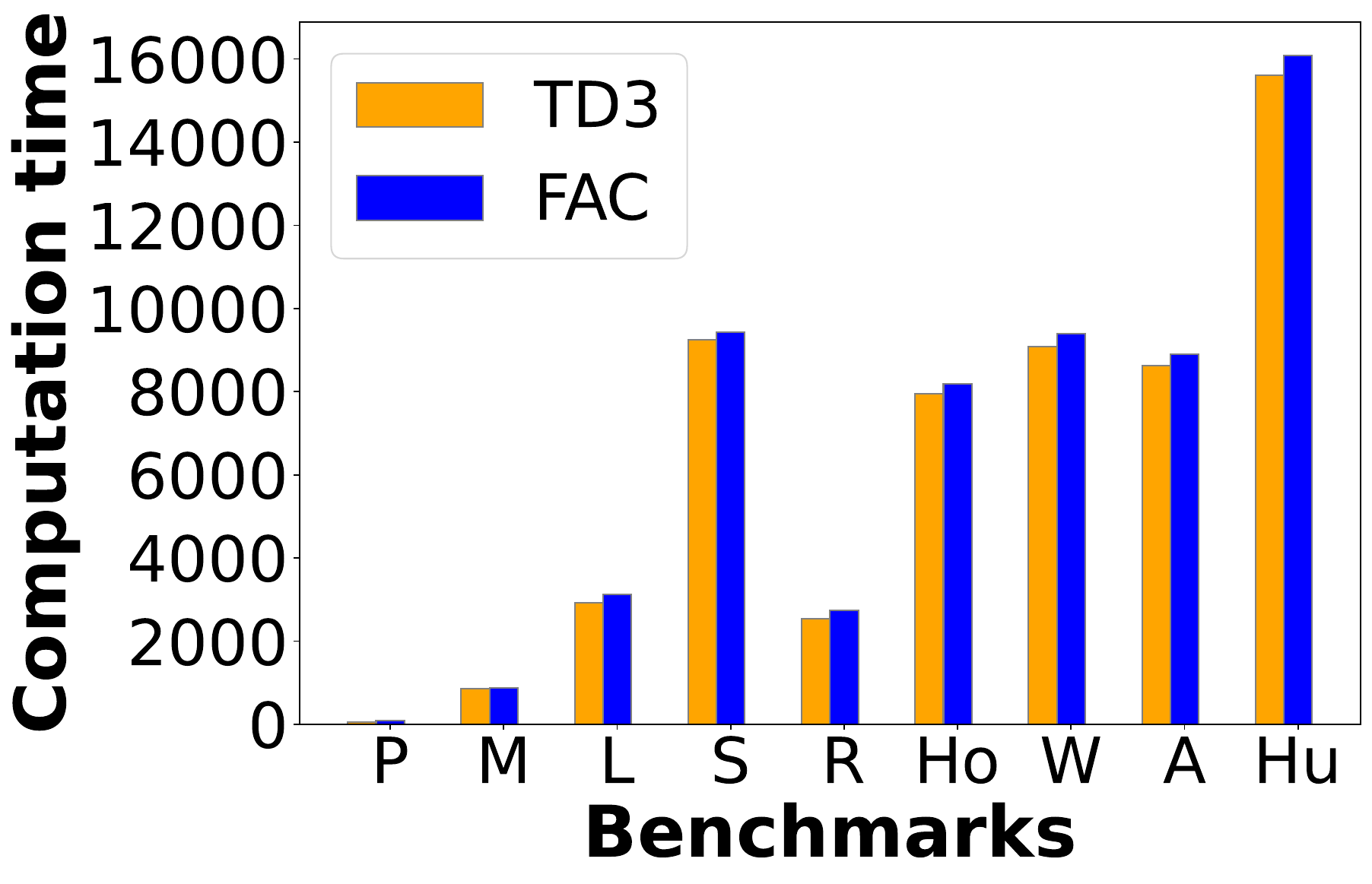}
         \caption{\textsf{FAC v/s TD3}}
         \label{fig:ctt}
    \end{subfigure}
     \caption{Comparison of computation time (in seconds) required for training using  FAC w.r.t. SAC and TD3.}
    \label{fig:ct}   
\end{figure}

\smallskip
\noindent
\textbf{Computation Time.}
The computation time overhead during training due to the FAC algorithm is negligible compared to overall time (refer Figure~\ref{fig:ct}). For instance, in the case of Hopper, we observed that the time taken by SAC and FAC algorithms were $18065\si{\second}$ and $18254\si{\second}$, respectively.

\begin{table*}
  \centering
  \caption{Comparison of Convergence (CP) and Reward (R) of FAC with baseline LABER w.r.t. SAC and TD3}
  \label{table-laber-comp}
  \begin{tabular}{l|rrr|rrr|r}
    \toprule
    Model    & $CP_{L}$ & $CP_{F}$ & $\Delta^{CP}$  & $R_L$ & $R_F$ & $\Delta^R$  & $P$\\
    \midrule
      Pendulum (SAC)  & $13600$  & $17600$  &  $-$  & $-180.17_{\pm209.52}$ & $-144.66_{\pm88.53}$ &  $+24.54$  & $2.18$ \\
     Pendulum (TD3)  & $18400$ &  $18400$ &  $-$ & $-142.29_{\pm87.62}$ & $-136.89_{\pm85.41}$ & $+3.94$  & $1.47$ \\
     \midrule
         MC (SAC)  & $29635$  & $23089$ & $22.09$   & $-7.07_{\pm1.37}$  & $-7.09_{\pm0.41}$ & $-0.28$  & $4.99$  \\
     MC (TD3)  &  $71131$  & $33590$ & $52.77$  & $-9.75_{\pm0.05}$  & $-5.97_{\pm0.07}$ &  $+63.31$  & $16.33$\\
\midrule
     LLC (SAC) &  $423928$ & $215108$  & $49.25$ & $183.49_{\pm18.10}$ & $179.72_{\pm16.05}$ &  $-2.05$  & $1.16$ \\
     LLC (TD3) &  $299600$ & $263927$  & $11.90$  & $139.37_{\pm47.06}$ & $146.43_{\pm26.90}$ & $+5.06$  & $3.00$\\
\midrule
Swimmer (SAC) & $468000$  & $188000$ & $59.83$   & $1.25_{\pm6.00}$  & $328.11_{\pm1.98}$ &  $+26148.8$  & $1563.88$ \\
     Swimmer (TD3)   &  $796000$  & $928000$ & $-$  & $34.36_{\pm3.11}$  & $114.86_{\pm1.91}$ &  $+234.28$  & $54.54$ \\
\midrule
     Reacher (SAC)  &  $172800$  & $171600$ &  $0.69$  & $19.32_{\pm10.17}$ & $19.96_{\pm10.23}$  &  $+3.31$  & $2.04$ \\
     Reacher (TD3) & $249600$  & $243600$ & $2.40$   & $13.33_{\pm12.07}$ & $17.10_{\pm10.28}$  &  $+28.28$  & $2.54$  \\
\midrule
    Hopper (SAC) &  $952414$ & $577319$  & $39.38$   & $1844.07_{\pm256.41}$ & $3529.13_{\pm6.49}$  &  $+91.37$ & $6.59$ \\
    Hopper (TD3)  & $877102$ & $848361$  & $3.27$   & $3315.74_{\pm82.07}$ & $3269.26_{\pm562.92}$  &  $-1.40$ & $3.08$ \\
\midrule
    Walker (SAC) & $863761$  & $779446$ & $9.76$  & $4321.31_{\pm843.80}$  & $5386.14_{\pm24.75}$ & $+24.64$   & $1.79$\\
     Walker (TD3)  & $648102$  & $855112$ &   $-$ &  $3809.35_{\pm36.37}$  & $4875.14_{\pm57.47}$ &  $+27.97$  & $1.69$ \\
\midrule
    Ant (SAC)  & $885086$  & $782363$ &  $11.61$  &  $3443.06_{\pm770.65}$  & $5057.42_{\pm1022.72}$ &  $+46.88$ & $1.68$   \\
      Ant (TD3)  &   $924276$  & $676999$ &    $26.75$ & $3082.34_{\pm1518.43}$  & $5416.09_{\pm139.93}$ &  $+75.71$  &  $2.33$  \\
\midrule
     Humanoid (SAC) & $1822733$  & $1079479$ &  $40.77$ & $4838.66_{\pm1294.53}$  & $5479.23_{\pm786.85}$ &  $+13.23$   & $1.40$\\
      Humanoid (TD3)  & $1999903$  & $1320449$ &  $33.97$  & $412.38_{\pm30.78}$  & $5217.76_{\pm396.35}$ &  $+1165.27$  & $28.97$ \\
    \bottomrule
  \end{tabular}

\end{table*}

\smallskip
\noindent
\textbf{Comparison with prioritization-based methods.}
\label{laber-comp}
%
A major approach to achieving sample efficiency in RL algorithms is sample prioritization. Thus, we compare our work with the most recently published prioritization algorithm LABER[17]. PER [28] is the most well-cited algorithm that introduced sample-prioritization. We did not include comparison results with PER as LABER has already been shown to be superior to PER. 


We present the comparison of FAC with LABER in Table~\ref{table-laber-comp}.
We observe that FAC outperforms LABER in terms of both convergence and accumulated reward for most benchmarks while maintaining a significantly smaller size for the replay buffer (The replay buffer size $|\mathcal{R}|$ for LABER is the same as the baseline SAC/TD3).
Comparing the convergence of FAC w.r.t. LABER, we observe an improvement up to $59.83\%$ in the case of SAC and $52.77\%$ in the case of TD3. In the case of Swimmer and Walker for TD3, convergence is delayed on account of improvement in reward accumulation. In terms of reward accumulation, FAC performs much better than LABER in almost all cases. 
For the Swimmer benchmark with SAC and the Humanoid benchmark with TD3, LABER cannot find a controller with sufficient reward accumulation, leading to a poor controller. However, FAC becomes successful in finding controllers with excellent accumulated rewards.

\section{Related Work}

The first use of experience replay in machine learning was in~\cite{Lin92}, where training was improved by repeating rare experiences.
Of late, replay buffers have played a significant role in the breakthrough success of deep RL~\cite{mnih13,Mnih15,Hasselt16,Lillicrap15}.
An ideal replay strategy is expected to sample transitions in a way that maximizes learning by the agent. Two different types of methods have been proposed in the literature to achieve this. The first focuses on which experiences to add and keep in the replay buffer, and the second focuses on how to sample important experiences from the replay buffer.

There has been very little research in developing strategies to control the entries into the replay buffer.
The authors in ~\cite{bruin15,Bruin16} suggest that for simple continuous control tasks, the replay buffer should contain transitions that are not close to current policy to prevent getting trapped in local minima, and also, the best replay distribution is in between on-policy distribution and uniform distribution. Consequently, they maintain two buffers, one
for on-policy data and another for uniform data, but this involves a significant computational overhead. This approach works in cases where policy is updated fewer times and hence is not suitable for complex tasks where policy is updated for many iterations. 
In~\cite{Isele18}, the authors 
show that a good replay strategy should retain the earlier samples as well as maximize the state-space coverage for lifelong learning. 
Hence, for stable learning, it is extremely crucial to identify which experiences to keep or discard from the replay buffer, such
that sufficient coverage of the state-action space is enforced.  


Most of the recent works have focused on sampling important transitions from the replay buffers to achieve sample efficiency. 
By default, the transitions in a replay buffer are uniformly sampled. 
However, many other alternatives have been proposed to 
prioritize samples during the learning process, most notably PER~\cite{per}, which assigns TD errors as priorities to samples that are selected in a mini-batch. However, PER is computationally expensive and does not perform well in continuous environments, and leads to the synthesis of sub-optimal control policies~\cite{Fujimoto20,oh21, oh22, fu19}.  
The issue of outdated priorities and hyper-parameter sensitivity
of PER have been addressed in~\cite{lahire22}, where the authors propose importance sampling for estimating gradients.
In~\cite{sinha22}, the authors propose re-weighting of experiences based on their likelihood under the stationary distribution of the current policy to minimize the approximation errors of the value function.  
The work in~\cite{novati19a} is based on the idea that transitions collected on policy are more useful to train the current policy and hence enforce similarity between policy and transitions in the replay buffer.
In~\cite{Zha19}, a replay policy is learned in addition to the agent policy. A replay policy helps in sampling useful experiences. 
All these methods are orthogonal to our approach, though all of them can potentially get benefited by using FAC to maintain unique samples in a replay buffer and reduce its size.
Moreover, the prioritization based methods use some approximation of TD-errors as priority, which is problematic as it is biased towards outliers. FAC, on the other hand, uniformly samples from replay buffer and hence is not biased towards outliers.

\section{Conclusion}

We have proposed a novel off-policy actor-critic RL algorithm for control policy synthesis for model-free complex dynamical systems with continuous state and action spaces. The major benefit of the proposed method is that it stores unique samples in the replay buffer, thus reducing the size of the buffer and improving the IID characteristics of the samples. We provide a theoretical guarantee for its superior convergence with respect to the general off-policy actor-critic RL algorithms. We also demonstrate experimentally that our technique, in general, converges faster, leads to a significant reduction in the size of the replay buffer and superior accumulated rewards, and is more sample efficient in comparison to the current state-of-the-art algorithms.
Going forward, we plan to explore the possibility of exploiting the capability of FAC in reducing the size of the replay buffer in synthesizing a control policy represented by a smaller actor network without compromising its performance.



\begin{acks}
The authors would like to thank anonymous reviewers for their valuable feedback. The first author acknowledges the Prime Minister’s Research Fellowship from the Government of India and the second author acknowledges the research grant MTR/2019/000257 from Science and Engineering Research Board of Government of India for supporting this research.
\end{acks}
\bibliographystyle{ACM-Reference-Format} 
\bibliography{references}



\end{document}